\documentclass[journal]{IEEEtran}
\usepackage{graphicx}
\usepackage{amsmath}
\usepackage{amssymb}
\usepackage{booktabs}

\usepackage{makecell}
\usepackage{amsfonts}
\usepackage[table]{xcolor}
\usepackage{tcolorbox}
\usepackage{multirow}
\usepackage{algorithm}
\usepackage{algorithmic}
\usepackage{pifont}

\ifCLASSINFOpdf
\else
\fi

\begin{document}
%
\title{Visual Position Prompt for MLLM based Visual Grounding}
%
%
%

\author{Wei~Tang$^\dagger$,
        Yanpeng~Sun,
        Qinying~Gu,
        and~Zechao~Li,~\IEEEmembership{Senior~Member,~IEEE}
\IEEEcompsocitemizethanks{
  \IEEEcompsocthanksitem W. Tang, Y. Sun and Z. Li are with School of Computer Science and Engineering, Nanjing University of Science and Technology, No.200 Xiaolingwei Road, Nanjing 210094, China. E-mail: \{weitang, zechao.li and yanpeng\_{sun}\}@njust.edu.cn (Corresponding authors: Zechao Li and Qinying Gu)
  \IEEEcompsocthanksitem Q. Gu are with Shanghai Artificial Intelligence Laboratory, No.129 Longwen Road, Shanghai 200232, China. E-mail: guqinying@pjlab.org.cn.
  \IEEEcompsocthanksitem {$^\dagger$} This work was done during his internship at Shanghai Artificial Intelliaence Laboratory.
  }
}

\markboth{Journal of \LaTeX\ Class Files,~Vol.~14, No.~8, August~2015}%
{Shell \MakeLowercase{\textit{et al.}}: Bare Demo of IEEEtran.cls for IEEE Journals}
%



\maketitle

\begin{abstract}
Although Multimodal Large Language Models (MLLMs) excel at various image-related tasks, they encounter challenges in precisely aligning coordinates with spatial information within images, particularly in position-aware tasks such as visual grounding. This limitation arises from two key factors. First, MLLMs lack explicit spatial references, making it difficult to associate textual descriptions with precise image locations. Second, their feature extraction processes prioritize global context over fine-grained spatial details, leading to weak localization capability. 
To address these issues, we introduce VPP-LLaVA, an MLLM enhanced with Visual Position Prompt (VPP) to improve its grounding capability. {VPP-LLaVA integrates two complementary mechanisms: the global VPP overlays a learnable, axis-like tensor onto the input image to provide structured spatial cues, while the local VPP incorporates position-aware queries to support fine-grained localization.}
To effectively train our model with spatial guidance, we further introduce VPP-SFT, a curated dataset of 0.6M high-quality visual grounding samples. Designed in a compact format, it enables efficient training and is significantly smaller than datasets used by other MLLMs (e.g., ~21M samples in MiniGPT-v2), yet still provides a strong performance boost. The resulting model, VPP-LLaVA, not only achieves state-of-the-art results on standard visual grounding benchmarks but also demonstrates strong zero-shot generalization to challenging unseen datasets.
The code and dataset are available at https://github.com/WayneTomas/VPP-LLaVA.
\end{abstract}


%
\IEEEpeerreviewmaketitle

\section{Introduction}
%
%
%
%
\begin{figure}[t]
    \centering
    \includegraphics[width=1\linewidth]{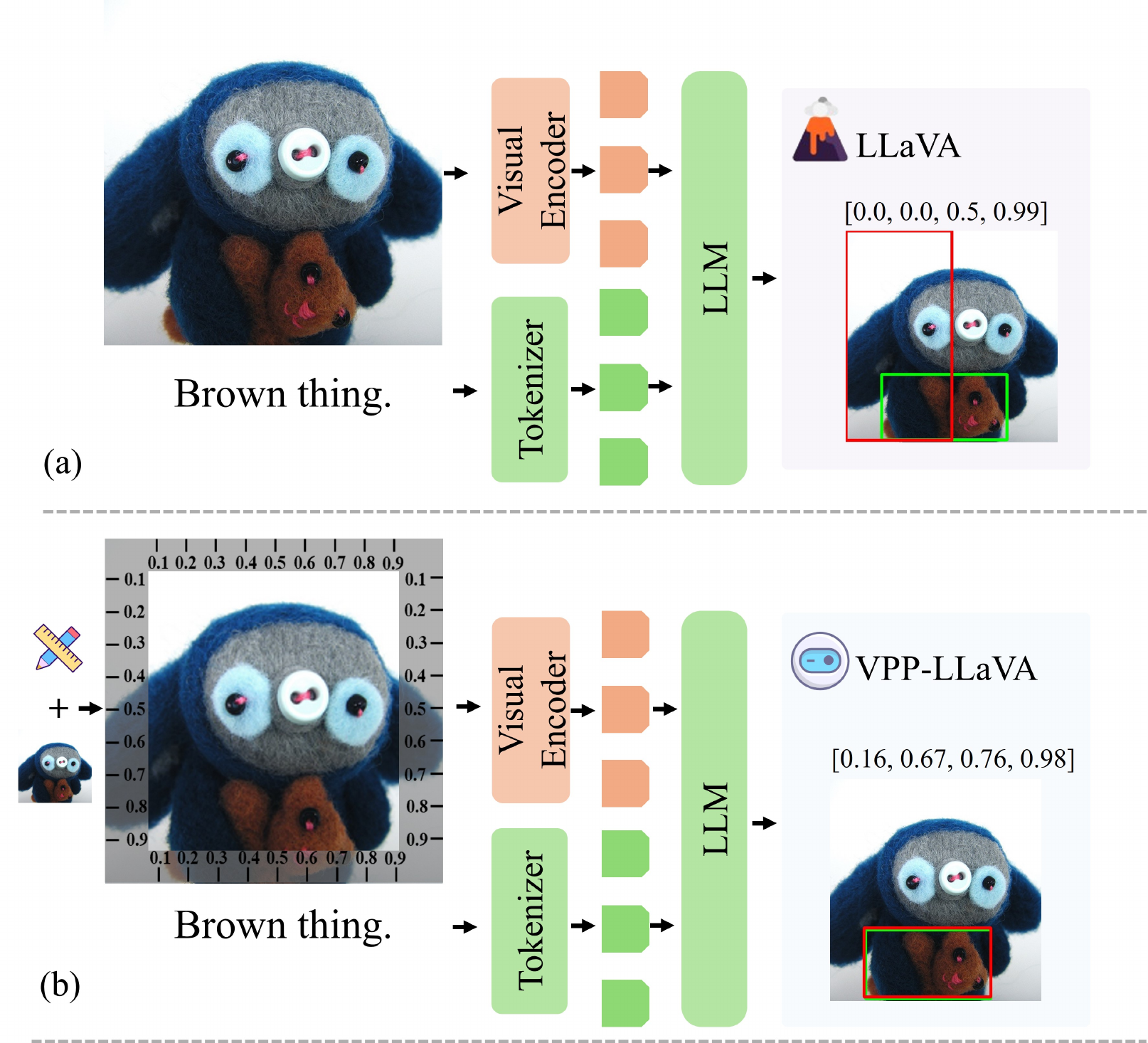}
    \caption{A visual grounding case study of MLLMs: (a) LLaVA-v1.5 outputs an inaccurate bounding box based on the given query expression. (b) When provided with a position reference, VPP-LLaVA produces a suitable result. The masked global VPP displays the coordinate regions using a semi-transparent gray overlay, indicating that this part is both visible and learnable. For brevity, some text instructions are omitted.}
    \vspace{-5mm}
    \label{fig:motivation}
\end{figure}

\begin{figure*}[t]
    \centering
    \includegraphics[width=1\linewidth]{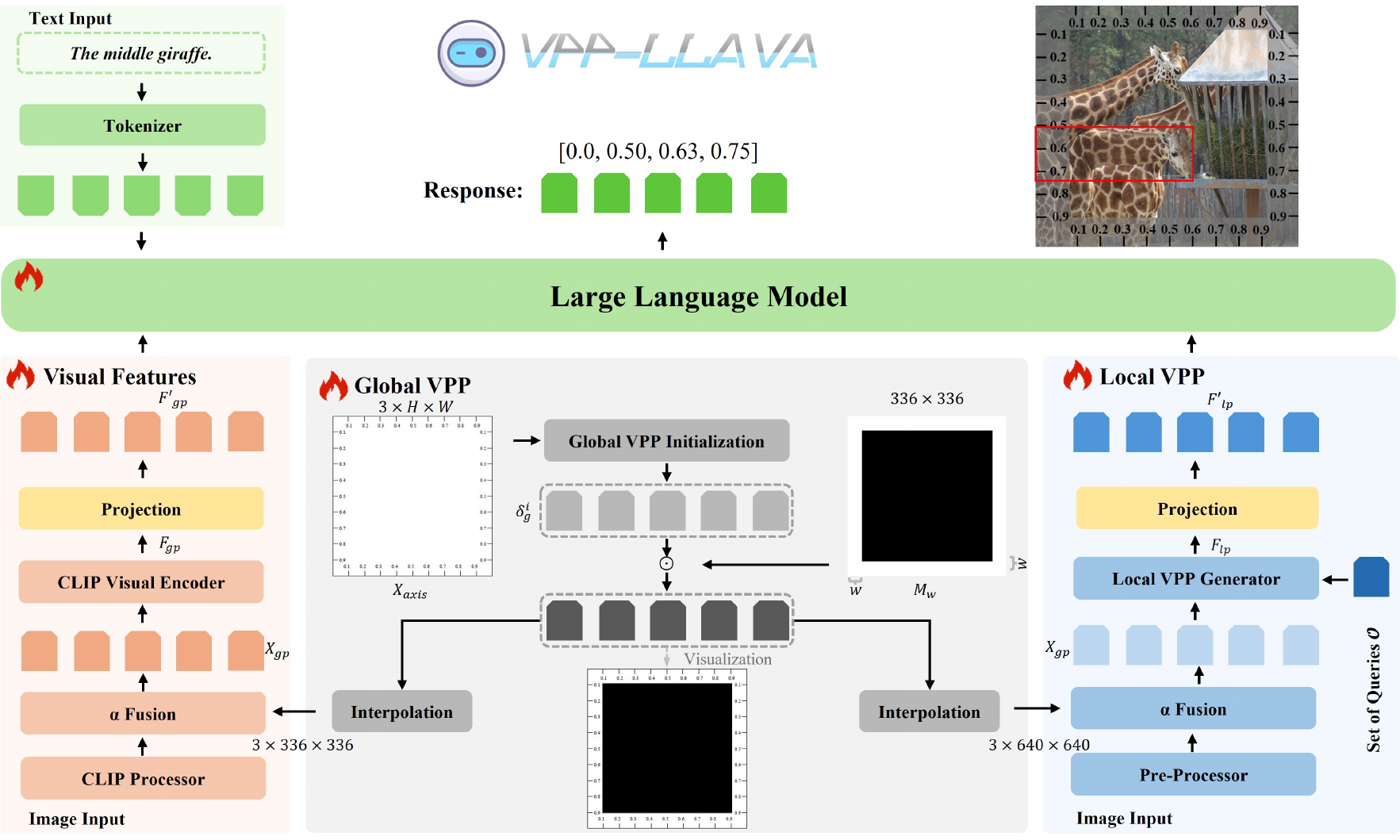}
    \caption{{An illustration of VPP-LLaVA, an MLLM-based visual grounding framework with Visual Position Prompt (VPP). We utilize the global VPP to provide a global position reference for MLLMs with foundational spatial cues. Additionally, a local VPP, serving as a local position reference, is introduced to further enhance and incorporate object spatial information. In the illustrative results, the coordinate regions of the masked global VPP are highlighted with a semi-transparent gray overlay, indicating that these areas are both visible and learnable. For brevity, some text instructions are omitted.}}
    \label{fig:framework}
\end{figure*}

\IEEEPARstart{M}{ulti-modal} Large Language Models (MLLMs) \cite{tmm_4,llava,llava1.5,llavanext,sun_1} achieve impressive results across various image-related tasks \cite{groma, ferret, Cong_1,tu_1}, 
earning considerable attention from the research community. Among these tasks, visual grounding—specifically Referring Expression Comprehension (REC)—stands out as a critical challenge \cite{ bu2025error,xie2025phrase,clip_vg,ferret,minigptv2}. Unlike pure detection tasks \cite{DAC, detr, anchor_detr}, visual grounding involves precisely identifying locations within an image based on free-form language expressions. It is fundamental for cognitive interactions between humans and machines, with applications such as image segmentation \cite{ctnet,singular}, remote sensing \cite{tgeo} and human-robot interaction \cite{tmm_7}.

While research \cite{shikra,cambrian-1,gpt4roi} indicates that MLLMs possess a reasonable ability for spatial understanding, there is a growing consensus that these models still require further enhancement, particularly for tasks involving precise spatial reasoning like visual grounding. As illustrated in Fig.~\ref{fig:motivation} (a), the grounding results from LLaVA-v1.5, for instance, reveal notable inaccuracies (the red box indicates the predicted bounding box, while the green box represents the ground truth). Although the predicted bounding box partially covers the target object, i.e., the brown toy, it suffers from both size and shape inaccuracies, failing to align well with the object’s true boundaries. These limitations highlight the need for more effective strategies to improve spatial alignment and object localization in MLLMs. To address these challenges, some studies are investigating the integration of advanced region-level enhancement modules and larger, more comprehensive visual grounding datasets into MLLMs \cite{minigptv2,qwen-vl,arcana,groma, groundingGPT}. Other approaches are exploring the incorporation of task-specific expertise, such as converting special tokens directly into bounding boxes with additional decoder, to improve localization accuracy \cite{lisa,llava_grounding}.

However, despite the use of the aforementioned methods to enhance the performance of MLLM in grounding tasks, studies suggest that MLLMs still face significant challenges in precisely aligning coordinates with spatial information in images \cite{dettoolchain,scaffolding}. One key issue lies in the models’ ability to effectively interpret and utilize spatial cues, which remain underutilized in many cases. As shown in Fig.~\ref{fig:motivation} (b), when we introduce a positional reference in the form of a coordinate axis, providing an explicit spatial guide, the model’s understanding of spatial relationships improves significantly. This reference allows the model to better interpret the spatial information of objects in the image, leading to more accurate localization. Specifically, the predicted bounding box becomes more aligned with the brown toy, reflecting the improved spatial reasoning and localization accuracy when positional references are incorporated.

Based on these observations, we propose integrating positional references as prompts within MLLMs to improve their visual grounding capabilities. While the coordinate axis shown in Fig.\ref{fig:motivation} (b) provides a global spatial guide, we further incorporate object position embeddings derived from detection models to supply local spatial cues by encoding object locations and semantic context. Such integration of global and local spatial priors has also been explored in other domains \cite{jiang2024global,jiang2024dvf}. These two types of Visual Position Prompt (VPP)—global and local—are complementary, with the global guide helping to establish overall spatial structure, and the local cues refining object-specific localization. 

Specifically, the global VPP is initialized in an axis-like form and overlaid onto the input image, providing a global spatial reference. This enables MLLMs to more effectively align coordinate information with spatial details across the image. To capture object-specific spatial and semantic information, we introduce a local VPP, which identifies potential objects within the image. This local reference helps the decoder integrate object-level details with other features. The combination of global and local VPP enables our model to better capture spatial information. Trained solely on our compact 0.6M-sample VPP-SFT dataset, which consolidates high-quality grounding annotations, the model achieves state-of-the-art performance. 
{Furthermore, VPP-LLaVA exhibits strong zero-shot generalization on challenging unseen datasets, particularly on the newly proposed multi-granular visual grounding benchmark GSEval-BBox~\cite{gseval}, which includes difficult samples such as part-object and multi-object grounding.}
Compared to other MLLMs such as MiniGPT-v2~\cite{minigptv2} and Qwen series~\cite{qwen2.5}, which rely on approximately 21M grounding samples, our method achieves superior performance owing to both the VPP design and the high-quality VPP-SFT dataset.

In summary, our contributions are shown as follows:
\begin{itemize}
    \item We propose VPP-LLaVA, an MLLM-based method for visual grounding with a Visual Position Prompt, along with a high-quality grounding instruction tuning dataset, VPP-SFT, which contains approximately 0.6M samples.
    \item We introduce novel global and local Visual Position Prompts that enable MLLMs to more effectively link spatial information within images to coordinate details, thereby enhancing their visual grounding capabilities.
    \item {Extensive experiments demonstrate that VPP-LLaVA achieves state-of-the-art performance and strong zero-shot generalization with far less training data than others, demonstrating its efficiency and effective design.}
\end{itemize}

\section{Related Work}
\subsection{Visual Grounding}
Visual grounding is a fundamental vision-language task that locates objects in an image based on a given free-formed linguistic expression \cite{tcsvt_1}. 

\noindent\textbf{Conventional methods.} 
Conventional visual grounding methods are typically divided into three categories \cite{transcp}: two-stage methods \cite{EARN,mattnet,2023cycleREC}, one-stage methods \cite{FAOA,tip_1}, and transformer-based methods \cite{VLTVG,transcp,MDETR,clip_vg,ofa,uninext}.
In pioneering two-stage methods such as EARN \cite{EARN} and MattNet \cite{mattnet}, a standard detection network generates region proposals, and the proposal that best matches the language query is selected through cross-modal matching. Due to the limitations of pre-trained detectors, one-stage methods are increasingly being adopted as an alternative solution. For example, FAOA, which is based on YOLOv3, is presented in \cite{FAOA}. it generates the bounding box using a YOLO-like network based on the fused features from visual and linguistic modalities.
With the advancement of Vision-Language Pre-trained (VLP) models, transformer-based methods are becoming increasingly popular for their strong performance and independence from pre-defined anchors, which is typical of one-stage methods. TransVG \cite{Transvg}, VLTVG \cite{VLTVG}, TransCP \cite{transcp}, MDETR \cite{MDETR}, CLIP-VG \cite{clip_vg}, OFA \cite{ofa}, and UNINEXT-L \cite{uninext} are among the most representative methods. These methods directly fuse visual and linguistic features using a transformer and model visual grounding as a regression task.
Despite the significant progress made by conventional methods in the field of visual grounding, these approaches are often specifically designed for individual tasks. Recently, researchers have shifted their focus toward Multimodal Large Language Models (MLLMs), leveraging their ability to unify visual grounding with other vision-related tasks. This integration streamlines task handling within a single, versatile model.

\noindent\textbf{MLLM-based methods.} 
With the tremendous success of LLMs in the field of natural language processing \cite{llama,vicuna}, researchers are considering expanding their application to the multimodal domain \cite{llava,llava1.5}. LLaVA \cite{llava} and MiniGPT-4 \cite{minigpt} are typical examples of such efforts. However, they still have room for improvement in the field of visual grounding. To enhance MLLMs' grounding capability, researchers are proposing the incorporation of more visual grounding data and advanced position-aware modules. For example, Shikra \cite{shikra}, Kosmos \cite{kosmos}, and MiniGPT-v2 \cite{minigptv2} discrete coordinates into specialized tokens to better align with MLLMs' input formats. Ferret \cite{ferret} introduces a Spatial-Aware Visual Sampler, while PINK \cite{pink} employs a self-consistent bootstrapping approach to enhance region sensitivity.
These methods still adhere to the standard MLLM framework, where the decoder of the MLLM directly outputs the coordinates. Since LLMs are inherently not well-suited for dense prediction tasks, another type of MLLM-based grounding method exists, which focuses on introducing a task-specific decoder.
For example, LISA \cite{lisa} and GLaMM \cite{glamm} leverage the SAM decoder for pixel-level reference image segmentation, while LLaVA grounding \cite{llava_grounding} introduces an additional grounding module to predict bounding boxes. 
Although they have achieved great progress in visual grounding, they still face challenges in precisely aligning coordinates with spatial information in images. In addition, almost all of these models are trained on large-scale datasets \cite{minigptv2, groundingGPT, groma, shikra}. For example, MiniGPT-v2 \cite{minigptv2} trains on approximately 21M grounding data samples. In contrast, our approach explicitly aids MLLMs in learning these correspondences by providing clear positional references, achieving state-of-the-art performance while requiring training on only a much smaller dataset.

\subsection{Visual Prompt}
Visual prompts are used by researchers to guide models in specific tasks \cite{tmm_7,interactive_seg_1,tnnls_1,jiang2024delving,visual_incontext}, as they are more direct than text prompts \cite{tcsvt_2,maple}. 
In the Vision-and-Language Pre-training (VLP) models, VPT \cite{vpt}, MaPLe \cite{maple}, and CMPA \cite{cmpa} add several learnable tokens before the CLIP visual embeddings, transferring CLIP to few-shot classification tasks. PEVL \cite{pevl} and CPT \cite{CPT} adapt position-sensitive vision-language tasks like visual grounding to mask token prediction. Recently, visual prompts are introduced in large vision models like SAM \cite{sam} and its variants \cite{vrp_sam,fine_grained_vp} to guide image segmentation. In the field of MLLMs, many works consider the visual prompt approach for specific downstream tasks. \cite{scaffolding} and \cite{dettoolchain} present position-guided visual prompts for GPT-4V, which highly rely on the model's own Chain-of-Thought ability. Ferret \cite{ferret} and ViP-LLaVA \cite{vip-llava} introduce hand-drawn free-form visual prompts, such as scribbles and arrows, for building user-friendly MLLMs. \cite{tvp} proposes a transferable visual prompt method that can be applied to different models on downstream tasks after training on a single model. \cite{rethink_som} suggest incorporating external knowledge, such as segmentation masks as visual prompts for MLLMs, to enhance their visual understanding performance.

\section{Methods}

\subsection{Overview}
In this section, we present VPP-LLaVA, an MLLM-based method with the Visual Position Prompt for visual grounding. As depicted in Fig.~\ref{fig:framework}, our overall framework follows that of LLaVA-v1.5 \cite{llava1.5}. It includes a CLIP-L/336 visual encoder for extracting image features, a 2-layer MLP for mapping these features to the LLM's feature space, and a Vicuna-v1.5 \cite{vicuna} model as the LLM backbone to process the tokenized text prompts. To help MLLMs precisely align coordinates and spatial information within images, we introduce both global and local Visual Position Prompts.

\subsection{Global Visual Position Prompt}\label{sec:GVPP}
\begin{figure}[t]
    \centering
    \includegraphics[width=0.9\linewidth]{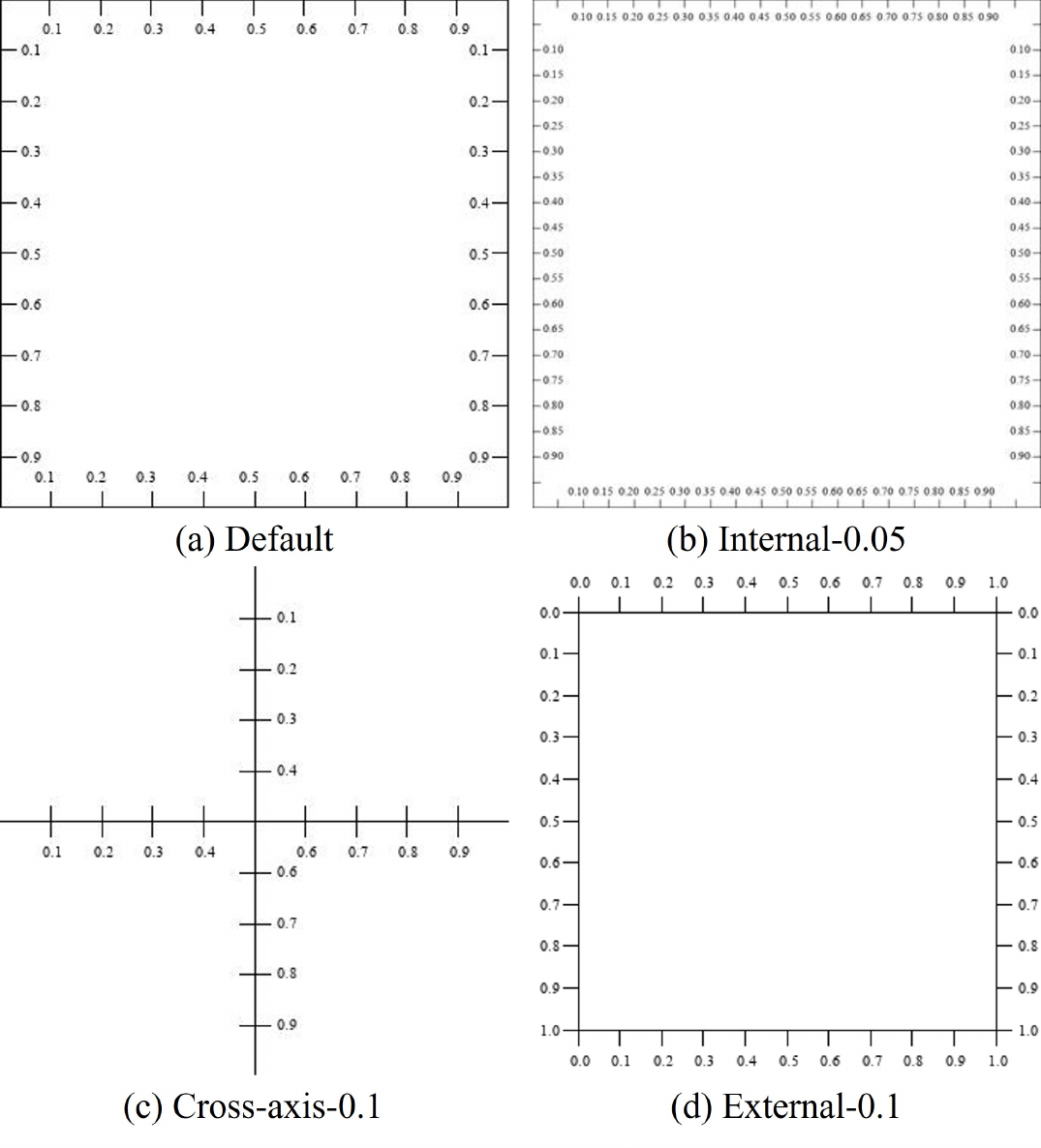}
    \caption{{Axis-like images used for initializing the global VPP. (a) Default version adopted in all main experiments; (b) Internal-0.05; (c) Cross-axis-0.1; (d) External-0.1 — the latter three are variants used in the ablation study to assess the effect of different positional priors. As these are the original resolution images used in training, some text elements may appear small in print; we recommend zooming in to view finer details.}}
    \label{fig:axis}
\end{figure}
Inspired by \cite{dettoolchain,tvp,scaffolding}, we introduce a learnable global Visual Position Prompt (VPP) $\delta_g \in \mathbb{R}^{3 \times 336 \times 336}$ to provide a global position reference for the MLLM. 
{It is initialized in an axis-like form and optimized without explicit constraints, enabling flexible adaptation during training.}
{
 \begin{equation}
    \begin{aligned}
        \delta_g^{i} \leftarrow \mathcal{T}_v(X_{axis}) \in \mathbb{R}^{3 \times 336 \times 336} ,
     \end{aligned}
\end{equation}
where $\delta_g^{i}$ is the initialized global VPP and $X_{axis}$ denotes the axis-like image. $\mathcal{T}v(\cdot)$ represents the visual encoder's preprocessing transformations, such as resize, padding and normalization in CLIP.}
{As illustrated in Fig.~\ref{fig:axis}, this design is consistent with the coordinate data format found in LLaVA’s training set, thereby providing the MLLM with a global positional reference. The axis-like RGB image used for initialization is a 3-channel image with a white background, while the axes, coordinate numbers, and tick marks are rendered in black. By default, we set a unit scale of 0.1 with axes along the edges (Fig.~\ref{fig:axis} (a)). Additional variants of the axis-like initialization image and the font size of the coordinate numbers are explored through detailed ablation studies presented in Section~\ref{sec:var_gvpp}.}

The global VPP is overlaid onto input images with standard transformations, such as resizing and padding. Mathematically, the process of the global VPP can be described as follows:
 \begin{equation}
 \label{eq:gvpp}
    \begin{aligned}
    X_{gp} = \alpha \cdot \mathcal{T}_v(X) + (1-\alpha) \cdot \mathcal{T}_{ipt}(\delta_g^{i} \odot M_w),
     \end{aligned}
\end{equation}
where $X$ is the input image. $\alpha$ is a trade-off parameter that controls the strength of the overlaid global VPP. $\mathcal{T}_{ipt}$ denotes an interpolation operation that scales the global prompt to fit the size of the processed input. 
{$M_w$ is a binary mask with a width of $w$ applied around the edges, which is used to control the visible region of the global VPP. Since the global VPP is initialized as an axis-like image with coordinate information along its edges, the masking mechanism is necessary because not all pixels need to be visible. As shown in Fig.~\ref{fig:framework} (global VPP module), the edges of the binary mask are set to 1, while the rest of the mask is set to 0, making the coordinates of the global VPP visible. Additionally, from the response results of VPP-LLaVA, we can see that the global VPP is marked with a semi-transparent gray overlay, indicating that this part is both visible and learnable.}

After we obtain the input image with the added global VPP $X_{gp}$, we send it to the visual encoder of LLaVA, i.e., CLIP-L/336, to get the image features.
 \begin{equation}
    \begin{aligned}
    F_{gp} = CLIP(X_{gp}).
     \end{aligned}
\end{equation}
Then, a two-layer MLP, similar to that used in LLaVA, is used to project the image features into the language space of the LLM.
 \begin{equation}
    \begin{aligned}
    F_{gp}' = MLP(F_{gp}).
     \end{aligned}
\end{equation}

 \subsection{Local Visual Position Prompt}\label{sec:LVPP}
To further enhance and incorporate object spatial information, we introduce a local VPP, which serves as a local position reference.


Specifically, in our implementation, we use DETR \cite{detr} as our local VPP generator to handle the entire local prompt generation process. Let $\mathcal{O}=\{o_{i}\}, i=1...k$ denote a set of object queries. By feeding the image with the overlaid global prompt and object queries into DETR's transformer encoder and decoder, we obtain 100 object query embeddings per image, capturing both potential object locations and semantic information.
 \begin{equation}
    \begin{aligned}
    F_{lp} = DETR(X_{gp}, \mathcal{O}),
     \end{aligned}
\end{equation}
where $F_{lp}$ denotes the generated local Visual Position Prompt, and $DETR(\cdot)$ denotes the DETR model, which includes a ResNet-101 backbone, a 6-layer transformer encoder, and a 6-layer transformer decoder.
Following LLaVA-v1.5, we also use a 2-layer MLP to map the features of the local prompt into the LLM's space.
 \begin{equation}
    \begin{aligned}
    F_{lp}' = MLP(F_{lp}).
     \end{aligned}
\end{equation}

The proposed local VPP distinguishes itself from other DETR-based methods in two key aspects: First, unlike existing two-stage visual grounding MLLMs that heavily rely on pre-trained detectors to extract bounding box proposals \cite{groma, glamm}, we dynamically generate object position embeddings that capture both spatial and semantic information. This allows the LLM decoder to integrate object-level details more effectively. Second, unlike methods such as ContextDET \cite{wang2024contextdet}, which incorporate DETR and add an extra box decoder while using the LLM solely for fusion, our method is a pure MLLM. This design not only simplifies the architecture but also enhances performance by enabling a more seamless integration of visual and language modalities.

\subsection{Instruction Tuning and Data Construction}
\begin{table*}[ht]
    \centering
    \caption{{Original annotation examples from four data sources. We randomly select two samples from each source for illustration. It can be seen that the formats of the four data sources are all different from each other. We unify them into the format of LLaVA-665K, and after refining and merging, we generate our VPP-SFT dataset.}}
    \label{tab:source_temp}
    \resizebox{\linewidth}{!}{
    \begin{tabular}{p{2.0cm} | l}
    \toprule
    Source (subset) & Annotations \\
    \midrule
    \multirow{4}{3cm}{LLaVA-665K} 
    & 1. [\{``from": ``human", ``value": ``\textless image\textgreater\textbackslash nPlease provide a short description for this region: [0.52, 0.59, 0.82, 0.83]."\}, \\
    &\quad \ \{``from": ``gpt", ``value": ``The giant doughnut with white icing and red , white , and blue sprinkles."\}] \\
    & 2. \{[``from": ``human", ``value": ``Please provide the bounding box coordinate of the region this sentence describes: white frosting."\}, \\
    &\quad \ \{``from": ``gpt", ``value": "[0.52, 0.59, 0.82, 0.83]"\}] \\
    \midrule
    \multirow{4}{3cm}{CB-GRD} 
    & 1. [\{``from": ``human", ``value": "Could you please tell me where is the black pants? "\}, \\
    &\quad \ \{``from": ``gpt", ``value": "I have provided the box of the black pants. \textless black pants:[245, 384, 283, 502]\textgreater"\}] \\
    & 2. \{[``from": ``human", ``value": "Draw the red and hanging and blue jacket out. "\}, \\
    &\quad \ \{"from: ``gpt", ``value": ``I have provided the box of the red and hanging and blue jacket. \textless red and hanging and blue jacket:[532, 354, 678, 570]\textgreater"\}] \\
    \midrule
    \multirow{4}{3cm}{CB-REF} 
    & 1. [\{``from": ``human", ``value": ``What can you see in this \textless region\textgreater ? \textless black charger:[354, 435, 451, 561]\textgreater"\}, \\
    &\quad \ \{``from": ``gpt", ``value": ``It is a black charger."\}] \\
    & 2. \{[``from": ``human", ``value": ``What is \textless it\textgreater ? \textless black clothing:[350, 112, 521, 153]\textgreater"\}, \\
    &\quad \ \{``from: ``gpt", ``value": ``It is a black clothing."\}] \\
    \midrule
    \multirow{2}{3cm}{Genixer}
    & 1. \ \ ``bbox": [75, 13, 240, 67], ``expression": ``the watch has a black strap" \\
    & 2. \ \ ``bbox": [313, 183, 431, 308], ``expression": ``a road through the grass" \\
    \bottomrule
    \end{tabular}
}
\end{table*}

\begin{figure}[t]
    \centering
    \includegraphics[width=0.8\linewidth]{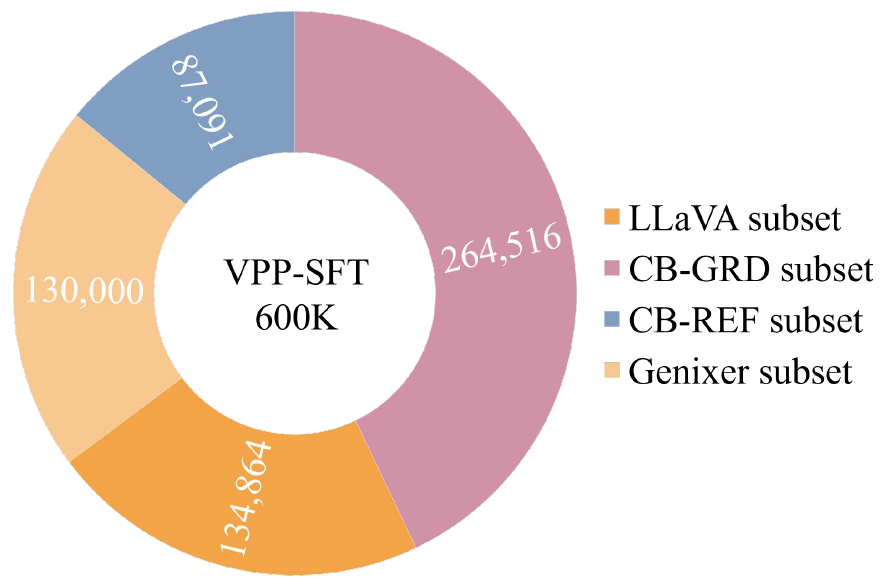}
    \caption{An overview of our constructed dataset VPP-SFT.}
    \label{fig:data_statistic}
\end{figure}

\begin{figure}[t]
    \centering
    \includegraphics[width=1\linewidth]{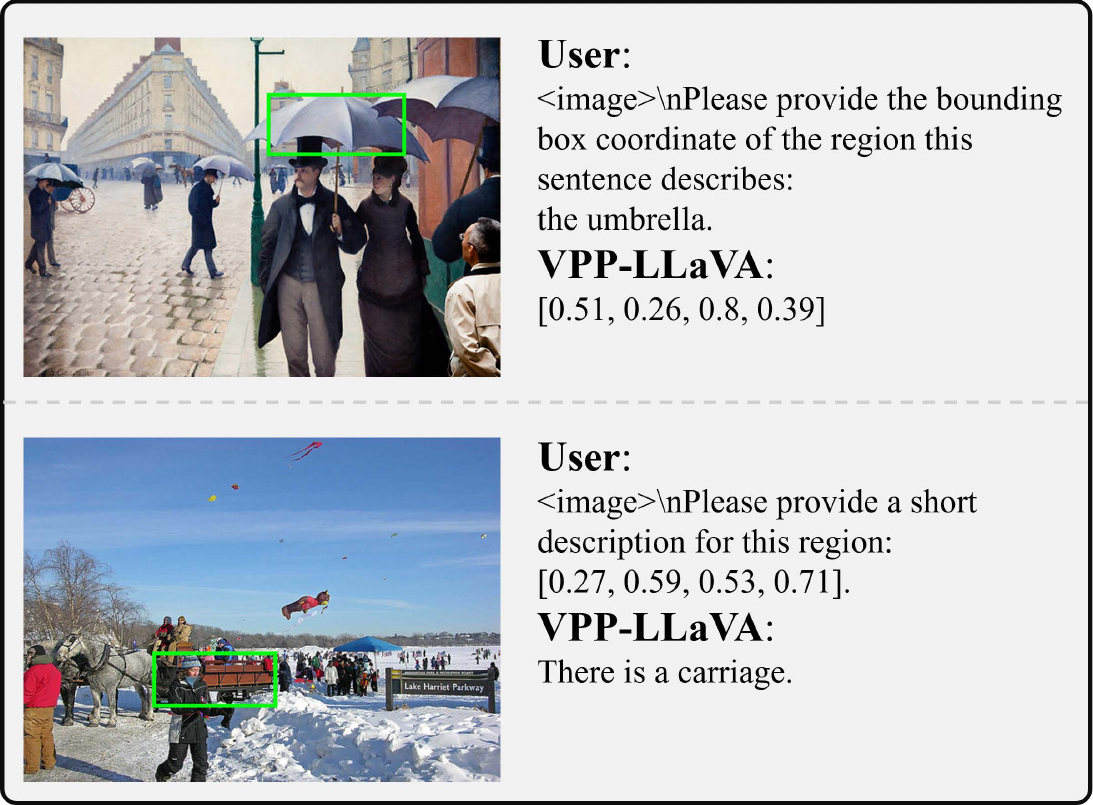}
    \caption{Training data examples in the VPP-SFT dataset. We provide two samples: the first is a training data sample for visual grounding, and the second is a region captioning sample used to maintain the language capabilities of the MLLM during downstream SFT. The green bounding boxes marked in the images are the Ground Truth.}
    \vspace{-5mm}
    \label{fig:data_example}
\end{figure}

Once the image features from both prompts are obtained, we directly concatenate them and feed the combined features, along with the query text instructions, into the LLM for processing. We utilize the pre-trained Vicuna-v1.5 \cite{vicuna} from LLaVA-v1.5 as our LLM. The response is generated by the LLM through autoregressive language modeling, i.e., by maximizing the likelihood of the next token prediction, which can be formulated as follows:

\begin{equation}
P(X_a \mid F^{'}, X_{q})=\prod_{i=1}^L P_{\theta}(x_i \mid F^{'}, X_{q}, X_{a,<i}),
\end{equation}

\begin{equation}
F^{'} = concatenate([F_{gp}', F_{lp}']).
\label{eq:fusion}
\end{equation}
Here, $F^{'}$ represents the concatenated visual features obtained by applying both the global and local VPP. The symbol $\theta$ denotes the parameters of the LLM, $X_q$ stands for the query text instructions, $L$ refers to the sequence length of the answer $X_a$, and $X_{a,<i}$ encompasses all answer tokens that precede the current prediction token $x_i$, where $i$ indicates the step in the next word token prediction process.

For instruction tuning of our model, we construct a dataset, VPP-SFT, containing approximately 0.6M samples, significantly smaller than other datasets commonly used for grounding in many MLLMs. For example, MiniGPT-v2 \cite{minigptv2} uses 21M grounding samples, KOSMOS-2 \cite{kosmos} uses 20M, and even the relatively smaller Shikra \cite{shikra} utilizes 4M samples. 

{As shown in Fig.~\ref{fig:data_statistic}, our constructed dataset primarily derives from four main sources: a subset of LLaVA-665K \cite{llava1.5}, CB-GRD \cite{chatterbox}, CB-REF \cite{chatterbox}, and Genixer \cite{genixer}, containing 134,864, 264,516, 87,091, and 130,000 conversations, respectively, due to the high quality of their original data. 
For each source, we provide corresponding annotation examples in Table~\ref{tab:source_temp}. As evident, the data formats vary widely. To enable consistent and efficient reuse in our training, we standardize them by converting bounding box coordinates into the LLaVA format, padding images along their longer edge, and normalizing the $(x_1, y_1, x_2, y_2)$ coordinates accordingly. Specifically, for LLaVA-665K, we extract both region captions and visual grounding conversations. In the subset of CB-GRD and CB-REF, we eliminate their special formatting schemes, preserve only referring expressions, and convert absolute coordinates (derived from original image dimensions) to normalized coordinates compatible with padded square inputs required by our model. Similarly, for Genixer, we construct conversations following grounding templates used in the LLaVA-665K and normalize the coordinate annotations to align with the LLaVA format.}
To retain LLaVA's original language capabilities, we incorporate a small amount of region-captioning data (e.g., CB-REF subset), formatted with instructions like: \textit{Please provide a short description for this region: xxx.} We aim for the tuned MLLM not only to provide coordinates in response but also to recognize and interpret coordinates as part of the language queries. The combination of these two types of data enhances the model's spatial awareness of object locations in images, improving its ability to perceive spatial information. A sample from our dataset is shown in Fig.~\ref{fig:data_example}.

It is important to note that VPP-SFT is not just a simple fusion of other datasets. Many MLLM papers introduce fragmented datasets that are hard to reuse. VPP-SFT tries to bring these scattered visual grounding datasets together for MLLMs. Unlike other instruction-tuning datasets for grounding, which often have complex formats with lots of special position tokens, our data format is refreshingly straightforward. This simplicity makes it easier to create new datasets based on ours and allows for more direct fine-tuning of models to boost MLLM spatial awareness.  Furthermore, we observe in our experiments, particularly on the zero-shot GSEval-BBox benchmark (Section~\ref{sec:zero-shot}), that VPP-LLaVA trained on our collected VPP-SFT dataset demonstrates impressive generalization to challenging part-object and multi-object scenarios. These results highlight the high quality and strong reusability of our dataset.

\section{Experiments}\label{sec:experiment}
\subsection{Benchmarks} \label{imple_details}
\noindent\textbf{RefCOCO/RefCOCO+/RefCOCOg.}
The datasets used in this study are derived from the MSCOCO \cite{mscoco}. RefCOCO contains 50,000 objects across 19,994 images, accompanied by 142,209 expression queries, with each query averaging 3.61 words in length. It is officially divided into training, val, testA, and testB sets, containing 120,624, 10,834, 5,657, and 5,095 expressions, respectively. Split testA focuses on expressions describing people, while testB emphasizes objects other than people. RefCOCO+ features 141,564 expressions for 49,856 objects within 19,992 images. Each query has an average length of 3.53 words. Contrary to RefCOCO, RefCOCO+ prohibits the use of absolute location expressions like \textit{right}, and \textit{top}, thus emphasizing the visual attributes of the referenced objects. The dataset is officially partitioned into training, val, testA, and testB, with 120,191, 10,758, 5,726, and 4,889 expressions, respectively. RefCOCOg features 49,822 objects across 25,799 images with 95,010 expression queries averaging 8.43 words. This dataset is available in two versions: RefCOCOg-google \cite{refcoco_google} and RefCOCOg-umd \cite{refcoco_umd}. Following the previous studies \cite{minigptv2, ferret, lion}, we report our performance on the RefCOCOg-umd dataset.

\noindent\textbf{ReferIt.} To evaluate the zero-shot of our model's visual grounding capabilities to other datasets, we conduct zero-shot visual grounding experiments on the ReferIt dataset. This dataset consists of 20,000 annotated images and is notable for containing ambiguous queries (e.g., \textit{any} and \textit{whole}). Besides, some of these samples is labeled with inaccurate bounding boxes. Officially, it is split into three subsets for training, validation, and testing, which comprise 54,127, 5,842, and 60,103 queries, respectively. 

\noindent\textbf{GSEval-BBox.} 
{To further assess the zero-shot capabilities of our proposed method in more challenging scenarios beyond ReferIt, we evaluate it on the recently introduced GSEval-BBox~\cite{gseval} dataset. This dataset contains 3,715 samples and spans four fine-grained expression types: \textit{stuff}, \textit{part-object}, \textit{multi-object}, and \textit{single-object}. Among these, part-object and multi-object references are particularly challenging, requiring models to accurately localize object subregions or reason over multiple entities simultaneously. Moreover, the referring expressions in GSEval-BBox are generally longer and more complex than those in RefCOCOg, posing an additional challenge for visual grounding models.}

\subsection{Implementation Details} \label{imple_details}
\noindent\textbf{Inputs.} For the visual encoder, we follow the same preprocessing steps as LLaVA: padding each image based on its longer side and then applying the CLIP/L-336 processor, which resizes the padded image to $336 \times 336$ pixels while preserving the original aspect ratio. For the input to the local VPP generator (DETR-ResNet101 in our model), we resize and pad the images to $640 \times 640$ pixels without any additional data augmentations.

\noindent \textbf{Training details.} We initialize our model with pre-trained LLaVA-v1.5-7B/13B parameters and fine-tune it using the AdamW optimizer with a cosine annealing scheduler. The learning rates are set as follows: 2e-5 for the LLM (Vicuna-v1.5-7B/13B), 2e-4 for the global VPP, 2e-5 for the local VPP generator, and 2e-4 for the projector that maps the local VPP to the LLM feature space. Unless otherwise specified, our default configuration refers to the 7B model (VPP-LLaVA-7B). Following recent works such as Cambrian-1 \cite{cambrian-1}, Eagle \cite{eagle}, and LLaVA-NeXT \cite{llavanext}, which demonstrate that unfreezing the visual encoder is greatly beneficial to vision-centric tasks, we adopt the same strategy, setting the learning rate of the visual encoder to 2e-6. Moreover, following Ferret, we fine-tune our model on the collected VPP-SFT dataset for 3 epochs, and the global batch size is set to 64, which takes roughly 30 hours on $8\times$ NVIDIA A100 GPUs (80GB) for the whole training process. Additionally, the width of the binary mask $w$ and the tradeoff parameter $\alpha$ in Eq.~\ref{eq:gvpp} are set to 30 pixels and 0.95 by default, respectively, with more details available in the Section~\ref{hyper}.

\subsection{Main Results}
\label{main_results}
\definecolor{lightblue}{RGB}{240, 248, 255} 
\begin{table*}[t]
\caption{Comparisons on visual grounding benchmarks. Qwen-VL-7B and Lion-4B are marked with the symbol $\dagger$ because they leverage a much larger visual encoder (1.9B ViT-bigG \cite{vit_big} and 1.1B EVA-G \cite{eva}, respectively). The highest performance is marked in bold, and the second-highest performance is marked with an underline.}
  \centering
  \renewcommand{\arraystretch}{1.3}
  \small  
  \resizebox{1.0\linewidth}{!}{
  \setlength\tabcolsep{7pt}
  \begin{tabular}{lcccccccccc}
  \toprule  
         \multirow{2}{*}{Method} &  \multirow{2}{*}{Model type} &  \multirow{2}{*}{Grounding Data Scale}&\multicolumn{3}{c}{RefCOCO} & \multicolumn{3}{c}{RefCOCO+}& \multicolumn{2}{c}{RefCOCOg}\\
         &&&val&testA&testB&val&testA&testB&val&test \\
         \midrule
         OFA-L \cite{ofa} & \multirow{4}{*}{Specialist} & 10M & 79.96&83.67&76.39&68.29&76.00&61.75&67.57 &67.58\\
         TransCP \cite{transcp}  &  & - & 84.25 & 87.38 & 79.78 & 73.07& 78.05 & 63.35 &- & - \\
         MDETR~\cite{MDETR} & & - & 86.75 & 89.58 & 81.41 & 79.52 & 84.09 & 70.62 & 81.64 & 80.89\\
         UNINEXT-L \cite{uninext} &&-& 91.43& 93.73& 88.93& 83.09& 87.90& 76.15& 86.91& 87.48\\
         
         \midrule
         KOSMOS-2 \cite{kosmos} &\multirow{10}{*}{Generalist}& 20M & 52.32&57.42&47.26&45.48&50.73&42.24&60.57&61.65 \\
         Shikra-7B \cite{shikra} && 4M & 87.01&90.61&80.24&81.60&87.36&72.12&82.27&82.19 \\
         Shikra-7B+Genixer \cite{genixer} &&$\sim$4.4M&87.48& 91.05& 81.77& 81.89& 87.43& 73.14& 81.99& 83.15\\
         Ferret-7B \cite{ferret} && 8.7M & 87.49&91.35&82.45&80.78&87.38&73.14&83.93&84.76 \\
         GroundingGPT \cite{groundingGPT}  & & $\sim$2.5M&88.02 & 91.55 & 82.47 & 81.61 & 87.18 & 73.18 & 81.67 & 81.99 \\
         MiniGPT-v2-7B \cite{minigptv2} && $\sim$21M & 88.06&91.29&84.30&79.58&85.52&73.32&84.19&84.31 \\
         Qwen-VL-7B \cite{qwen-vl}$^\dagger$ && $\sim$21M &88.55 &92.27& 84.51 &82.82& 88.59& 76.79& \underline{85.96}& \underline{86.32}\\
         PINK \cite{pink} && 5M &88.70&92.10&84.00&81.80&88.20&73.90&83.90&84.30 \\
        Groma \cite{groma} &&$\sim$26M& \underline{89.53} & \underline{92.09} & \textbf{86.26} & \underline{83.90} & \underline{88.91} & \textbf{78.05}& \textbf{86.37} & \textbf{87.01} \\
        \rowcolor{lightblue} VPP-LLaVA-7B & & \textbf{$\sim$0.6M} & \textbf{90.37} &\textbf{92.89} &\underline{85.77} &\textbf{84.65} &\textbf{89.84} &\underline{76.99} &85.33 &85.52  \\
         
         \midrule
         Shikra-13B \cite{shikra} & \multirow{5}{*}{Generalist} & 4M & 87.83&91.11&81.81&82.89&87.79&74.41&82.64&83.16 \\
         Ferret-13B \cite{ferret} && 8.7M & 89.48& \underline{92.41} & 84.36&82.81&88.14&75.17&\underline{85.83}&\textbf{86.34} \\
         Griffon-v2-13B \cite{griffon_v2} && $\sim$13M & 89.60&91.80&\textbf{86.50}&81.90&85.50&76.20&\textbf{85.90}&86.00 \\
         Lion-12B \cite{lion}$^\dagger$ && 7.2M &89.80& \textbf{93.02} & 85.57 & \underline{83.95} & \underline{89.22} & \underline{78.06} & 85.52& 85.74\\
         \rowcolor{lightblue} VPP-LLaVA-13B & & \textbf{$\sim$0.6M} &\textbf{90.32} & \textbf{93.02} & \underline{86.34} & \textbf{84.65} & \textbf{90.78} & \textbf{79.06} & 85.64 & \underline{86.01}  \\
     \bottomrule
  \end{tabular}}
  
   \label{tab:main_results}
\end{table*}

In Table~\ref{tab:main_results}, we report the accuracy performance with an IoU threshold of 0.5 on widely used visual grounding benchmarks \cite{mscoco,refcoco_umd}: RefCOCO, RefCOCO+, and RefCOCOg. We categorize the compared state-of-the-art models into two types: specialist and generalist models. The methods marked with $\dagger$ indicate that they use ViT-BigG (1.9B) \cite{vit_big} and EVA-G (1.1B) \cite{eva} as their visual encoders, respectively, and are significantly larger than CLIP-L/336 used in our model. We also report the scale of the visual grounding data used for training the visual grounding capabilities of these models. From the results, we draw several observations as follows:

Firstly, as a generalist model, our VPP-LLaVA-7B demonstrates strong visual grounding capability, outperforming most of the specialist models. For instance, our method exceeds the performance of OFA-L, TransCP, and MDETR by a substantial margin. Even when compared to the most state-of-the-art specialist model, UNINEXT-L, which is trained on a diverse range of grounding and detection datasets such as object365 and the SOT\&VOS datasets, our model surpasses it on the three splits of RefCOCO+ with absolute margins of 2.56\%, 1.94\%, and 0.84\%, respectively.

Secondly, for the RefCOCO dataset, which involves short and position-based expressions averaging 3.6 words, our model achieves the best performance among the group of generalist models compared to others with over 4M grounding data samples. This demonstrates that the introduction of both global and local VPPs is highly effective. These prompts provide position references for MLLMs to locate objects, which facilitates the MLLM's learning and establishment of associations between spatial information and coordinates.

Thirdly, RefCOCO+, which excludes absolute location expressions such as \textit{right}, \textit{top}, and \textit{left}, significantly impacts the grounding achievements of most generalist models. In contrast, our method, leveraging both global and local VPPs, compensates for the lack of absolute location terms and achieves commendable grounding results.

Finally, RefCOCOg, which contains long and more complex queries, is an even more challenging dataset. Models with relatively small amounts of training data, such as Shikra and GroundingGPT, perform less well. Fortunately, despite differences in data scale and visual encoder capabilities, our method achieves competitive performance on RefCOCOg compared to current state-of-the-art methods such as Qwen-VL-7B and Groma. This phenomenon further demonstrates the effectiveness of our proposed method. By incorporating VPPs, our MLLM can more effectively align spatial information within images to coordinates, even when trained on significantly smaller datasets (0.6M vs. over 20M).

To sum up, the experimental results demonstrate the superiority and effectiveness of our VPP-LLaVA. 

\subsection{Scaling Up}
To perform a more comprehensive evaluation of the proposed model, we scaled up VPP-LLaVA to 13B settings. We use Vicuna-v1.5-13B, trained with LLaVA-v1.5-13B, as our LLM. The remaining settings are consistent with other configurations mentioned in the Section~\ref{main_results}. The results are shown in Table~\ref{tab:main_results}.

For the RefCOCO dataset, despite the limited scale of grounding data used for training, VPP-LLaVA-13B achieves either the best or comparable performance to the top-performing method across all three splits (val, testA, and testB). Moreover, The evaluation results on RefCOCO+ confirm the effectiveness of our approach, showing consistent performance improvements observed in previous experiments. Notably, VPP-LLaVA-13B outperforms Griffon-v2-13B by a significant margin. This improvement likely stems from Griffon-v2-13B's limitations in handling scenarios like RefCOCO+, where absolute positional terms are absent. In contrast, our Visual Position Prompts (VPPs) enable VPP-LLaVA-13B to infer positional information effectively, resulting in more accurate bounding box predictions. These findings align with earlier observations.

The results on RefCOCOg are also consistent with the previous experimental results obtained with 7B settings. Even with differences in the visual encoder and training scale, our model still achieves competitive performance against state-of-the-art methods, thereby proving the effectiveness of our proposed approach.

In summary, after scaling up to 13B parameters, VPP-LLaVA consistently maintains its advantages, proving that the enhancements we propose are both scalable and effective.

\subsection{Zero-Shot Generalization on Unseen Datasets}
\label{sec:zero-shot}
    \begin{table}[t]
    \caption{Zero-Shot Generalization Comparisons on ReferIt Dataset.}
    \definecolor{lightblue}{RGB}{240, 248, 255} 
      \centering
        \setlength{\tabcolsep}{20pt}
        \renewcommand{\arraystretch}{1.3}
        \resizebox{1.0\linewidth}{!}{
        \begin{tabular}{lcc}
        \toprule  
        Method  & ReferIt val & ReferIt test \\
        \hline
        LLaVA-v1.5-7B & 48.95 & 47.42 \\
        \rowcolor{lightblue}VPP-LLaVA-7B & \textbf{57.55} & \textbf{56.53} \\
        \bottomrule
      \end{tabular}
      }
      \label{tab:transfer_referit}
    \end{table}

    \definecolor{lightblue}{RGB}{240, 248, 255} 
    \begin{table}
    \caption{{Zero-shot generalization results on the newly proposed challenging GSEval-BBox benchmark. Our method achieves a significant performance lead, demonstrating strong generalization to unseen grounding scenarios.}}
    \centering
    \setlength{\tabcolsep}{3pt}
    \small
    \begin{tabular}{lccccc}
    \toprule
    \multirow{2}{*}{Method}  & \multicolumn{5}{c}{GSEval-BBox}  \\
             & Stuff & Part & Multi & Single & All\\ 
    \midrule
    InternVL2.5-78B~\cite{internvl2.5} & 85.3 & 16.8 & 63.2 & 55.7  & 62.2      \\
    InternVL2.5-8B~\cite{internvl2.5}  & \underline{91.3} & 7.3 & 65.7 & 47.3  & 58.2      \\
    Qwen2.5-VL-72B~\cite{qwen2.5}      & 88.4 & 31.2 & 42.8 & 64.6  & 62.5       \\
    Qwen2.5-VL-7B~\cite{qwen2.5}       & \textbf{93.0} & 17.6 & 75.9 & 59.1  & 66.7     \\
    DeepSeek-VL2~\cite{deepseekvl2}    & 86.5 & 12.7 & 64.8 & 51.2  & 60.8    \\
    Mistral-Small-3.1-24B~\cite{mistral} & 16.0 & 3.5 & 20.3 & 10.7  & 13.2 \\
    Ferret-7B~\cite{ferret}             & 82.1 & 17.6 & 56.0 & 43.2  & 53.3 \\
    Ferret-13B~\cite{ferret}            & 80.6 & 21.3 & 58.0 & 46.6  & 55.1 \\
    \rowcolor{lightblue} VPP-LLaVA-7B & 52.1 & \underline{67.9} & \underline{82.7} & \underline{68.7} & \underline{67.0} \\
    \rowcolor{lightblue} VPP-LLaVA-13B & 54.5 & \textbf{69.9} & \textbf{83.8} & \textbf{69.5} & \textbf{68.4} \\
    \bottomrule
    \end{tabular}
    \label{tab:gseval}
    \end{table}
{We evaluate the zero-shot generalization capability of VPP-LLaVA on two datasets not seen during training: ReferIt and the newly proposed GSEval-BBox. Since ReferIt has been included in the training set of many existing MLLMs, we report only the performance of the baseline LLaVA-v1.5-7B for fair comparison. As shown in Table~\ref{tab:transfer_referit}, VPP-LLaVA-7B outperforms the baseline by a large margin, with an absolute accuracy gain of 8.6\% on the validation split and 9.1\% on the test split.}

{To further assess the robustness and generalization of our approach in complex scenarios, we conduct a comprehensive zero-shot evaluation on GSEval-BBox~\cite{gseval}, a challenging benchmark featuring diverse expression types including stuff, part-object, multi-object, and single-object. As shown in Table~\ref{tab:gseval}, VPP-LLaVA-13B achieves 69.9\% accuracy on the part split, significantly outperforming the next best model (Qwen2.5-VL-72B) by a wide margin. On multi-object expressions, VPP-LLaVA-7B attains 82.7\%, clearly ahead of Qwen2.5-VL-7B. For single-object cases, our model also achieves the highest score of 69.5\%, setting a new state-of-the-art.}

{These results highlight the strong generalization ability of VPP-LLaVA in zero-shot visual grounding. We attribute this performance to the synergy between our proposed VPP design and the high-quality supervision provided by the VPP-SFT dataset, which enables the model to generalize well across diverse and challenging scenarios.}

\subsection{Transferability Studies for VPP}
\begin{table}[t]
\caption{Transferability Study of VPP on LLaVA-NeXT.}
\setlength\tabcolsep{2.2pt}
  \centering
  \renewcommand{\arraystretch}{1.3}
  \resizebox{1.0\linewidth}{!}{
  \begin{tabular}{lccccccccc}
  \toprule  
            \multirow{2}{*}{Method}&\multicolumn{3}{c}{RefCOCO} & \multicolumn{3}{c}{RefCOCO+}& \multicolumn{2}{c}{RefCOCOg}\\
            &val&testA&testB&val&testA&testB&val&test \\
            \hline
            LLaVA-NeXT & 84.74 & 89.67 & 77.29 & 77.27 & 85.71 & 66.86 & 80.12 & 79.68 \\
            \rowcolor{lightblue} LLaVA-NeXT+{VPP} & \textbf{90.28} & \textbf{93.26} & \textbf{86.73} & \textbf{84.75} & \textbf{90.48} & \textbf{77.99} & \textbf{85.60} & \textbf{85.53}  \\
     \bottomrule
  \end{tabular}
  }
  \label{tab:transfer_llava_next}
\end{table}




\begin{table}[t]
\caption{Comparisons of region captioning on RefCOCOg. Models marked in gray indicate the use of larger language models. The highest performance is marked in bold, and the second-highest performance is marked with an underline.}
\definecolor{lightblue}{RGB}{240, 248, 255} 
  \centering
    \renewcommand{\arraystretch}{1.3}
    \footnotesize
  
  \setlength{\tabcolsep}{3.4mm}{
    
    \begin{tabular}{lccc}
    \toprule
    Method  & METEOR & BERTScore-F1 & CIDER \\
    \hline
    GPT4ROI & 9.7 & \textbf{87.3} & - \\
    Kosmos-2 & \textbf{12.2} & \underline{87.1} & \underline{60.3}  \\
    LLaVA-v1.5-7B & 12.0  & 86.9 & \textbf{73.1} \\
    \textcolor{gray}{Griffon-v2-13B} & \textcolor{gray}{12.1} & \textcolor{gray}{-} & \textcolor{gray}{72.5}  \\
    \textcolor{gray}{ChatterBox-13B} & \textcolor{gray}{14.5} & \textcolor{gray}{88.0} & \textcolor{gray}{-}  \\
    \hline
    \rowcolor{lightblue}VPP-LLaVA-7B & \underline{12.1} & \underline{87.1} & \textbf{73.1}  \\
    \bottomrule
  \end{tabular}}
  \label{tab:region_cap}
\end{table}
{Transferability studies in our framework focus on two key aspects:
(1) whether the proposed VPP mechanism can be effectively transferred to MLLMs beyond LLaVA-v1.5; and (2) whether VPP-LLaVA preserves strong language generation capabilities.}

{For question (1), we apply VPP to LLaVA-NeXT-7B with the same training strategy, and the results in Table~\ref{tab:transfer_llava_next} show consistent improvements in visual grounding performance. This confirms that VPP effectively enhances grounding ability across different model architectures.}

For question (2), we evaluate the language generation capability of VPP-LLaVA on the coordinate-aware language generation task, i.e., region captioning, using the RefCOCOg dataset. As shown in Table~\ref{tab:region_cap}, VPP-LLaVA-7B achieves a CIDEr score of 73.1, comparable to or better than some larger models. Despite its smaller size and relatively limited training data, VPP-LLaVA maintains strong localization and captioning performance.


\subsection{Ablation Study} \label{ablation}

\begin{figure*}[!h]
    \centering
    \includegraphics[width=1\linewidth]{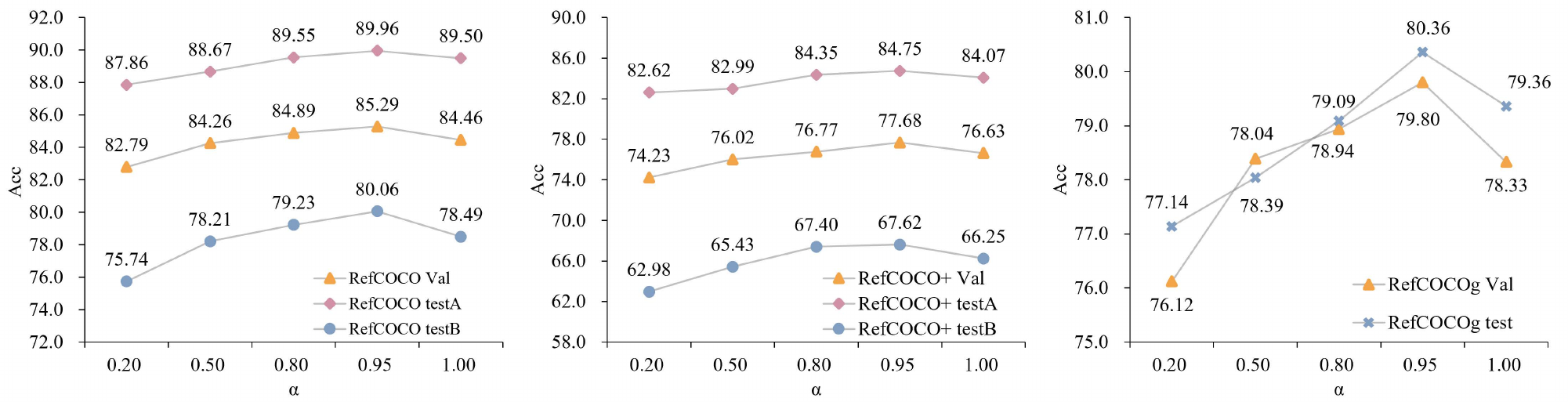}
    \caption{The ablation study of the trade-off parameter $\alpha$ in global VPP. The left, middle, and right figures present results on RefCOCO, RefCOCO+, and RefCOCOg, respectively. Overall, the performance indicates that $\alpha = 0.95$ achieves the best results.}
    \label{fig:abltion_alpha}
\end{figure*}

\begin{figure*}[!h]
    \centering
    \includegraphics[width=1\linewidth]{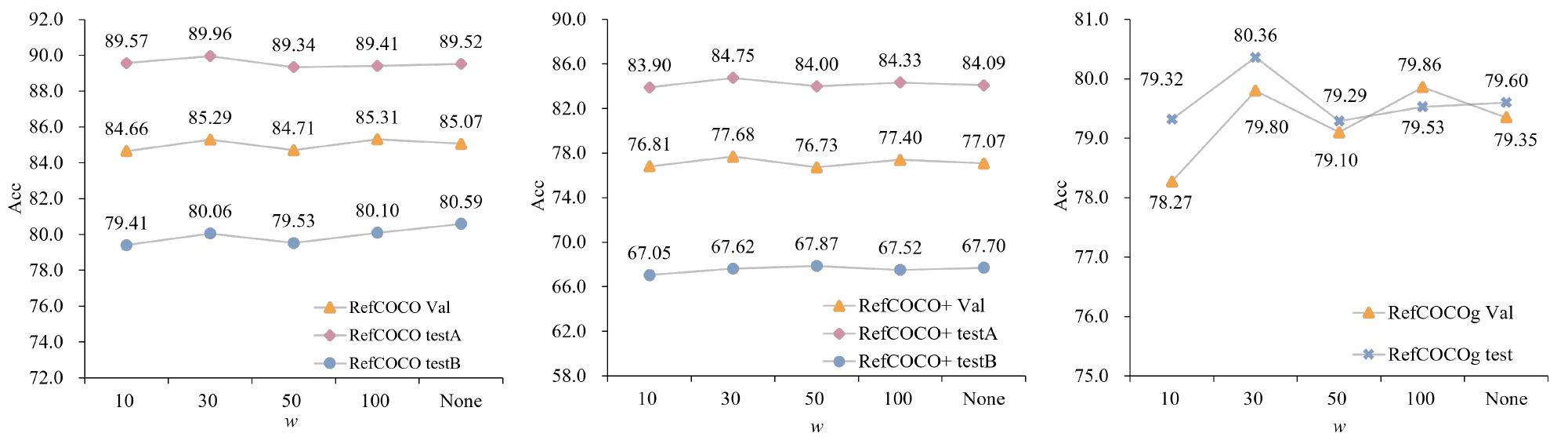}
    \caption{The ablation study of the width $w$ of the binary mask $M_w$ in global VPP. The left, middle, and right figures present results on RefCOCO, RefCOCO+, and RefCOCOg, respectively. Overall, the performance indicates that $M_w = 30$ pixels achieves the best results.}
    \label{fig:ablation_width}
\end{figure*}

\definecolor{lightblue}{RGB}{240, 248, 255} 
\begin{table*}[t]
\caption{The ablation study of different modules in our proposed method.}
\setlength\tabcolsep{2pt}
  \centering
  \renewcommand{\arraystretch}{1.3}
  \scalebox{1}{
  \begin{tabular}{cccccccccc}
  \toprule  
         \multirow{2}{*}{Global VPP} &   \multirow{2}{*}{Local VPP}&\multicolumn{3}{c}{RefCOCO} & \multicolumn{3}{c}{RefCOCO+}& \multicolumn{2}{c}{RefCOCOg}\\
         &&val&testA&testB&val&testA&testB&val&test \\
         \midrule
          &  & 84.58 & 89.62 & 78.07 & 77.04 & 83.79 & 67.00 & 78.23 & 78.63 \\
          \ding{51} & & 84.95 & 89.87 & 78.78 & 77.20 & 84.30 & 67.27 & 79.27 & 79.79 \\
           & \ding{51} & 84.46 & 89.50 & 78.49 & 76.63 & 84.07 & 66.25 & 78.33 & 79.36  \\
           
         \rowcolor{lightblue} \ding{51}  & \ding{51} & \textbf{85.29} &\textbf{89.96} &\textbf{80.06} &\textbf{77.68} & \textbf{84.75} & \textbf{67.62} & \textbf{79.80} & \textbf{80.36}  \\
         \rowcolor{lightblue} & & \textcolor{red}{(+0.71)} & \textcolor{red}{(+0.34)} & \textcolor{red}{(+1.99)} & \textcolor{red}{(+0.64)} & \textcolor{red}{(+0.96)} & \textcolor{red}{(+0.62)} & \textcolor{red}{(+1.57)} & \textcolor{red}{(+1.73)} \\
     \bottomrule
  \end{tabular}
  }
  
   \label{tab:ablation_components}
\end{table*}

To conserve time and computational resources, we freeze the visual encoder (except in Section~\ref{ablation:freeze}) and randomly select 150K samples from VPP-SFT to form our ablation training set. All subsequent ablation results are based on this subset and are specific to the VPP-LLaVA-7B model.

\subsubsection{Components}
Table \ref{tab:ablation_components} shows VPP-LLava's performance with different components on grounding benchmarks. Without the proposed VPP, the model degrades to LLaVA-v1.5 (the baseline, which is also trained on the same 150K subset with the same training strategy as described earlier for fairness). Adding the global VPP improves baseline performance across datasets, especially on RefCOCOg, with a 1.5\% gain, highlighting its effectiveness in providing a reliable spatial reference. The local VPP alone shows mixed results, sometimes comparable to or slightly below baseline, likely due to limited data hindering effective DETR feature alignment. However, combining global and local VPPs yields a synergistic effect, enhancing grounding accuracy by around 2\% on RefCOCO testB, as the global reference supports overall spatial orientation while the local VPP contributes region-specific semantic cues, jointly boosting performance.

\subsubsection{Impact of Trainable Global VPP}
\definecolor{lightblue}{RGB}{240, 248, 255} 
\begin{table}[t]
\caption{The ablation study of the impact about the global VPP being trainable or non-trainable.}
\setlength\tabcolsep{3pt}
  \centering
  \renewcommand{\arraystretch}{1.3}
  \small  
  \begin{tabular}{ccccccccc}
  \toprule  
            \multirow{2}{*}{Trainable}&\multicolumn{3}{c}{RefCOCO} & \multicolumn{3}{c}{RefCOCO+}& \multicolumn{2}{c}{RefCOCOg}\\
            &val&testA&testB&val&testA&testB&val&test \\
            \midrule
            \ding{55} & 84.77  & 89.57 & 79.27 & 77.04 & 83.43 & 67.11 & 79.04 & 79.88 \\
         \rowcolor{lightblue} \ding{51} & \textbf{85.29} &\textbf{89.96} &\textbf{80.06} &\textbf{77.68} & \textbf{84.75} & \textbf{67.62} & \textbf{79.80} & \textbf{80.36}  \\
     \bottomrule
  \end{tabular}
  
   \label{tab:ablation_freeze_vpt}
\end{table}

In our approach, the default setting allows the global VPP to be trainable. It is essential to examine the impact of training versus not training the global VPP. To conserve time and computational resources, we maintain the same setup as in our previous ablation study, using a 150K subset sampled from VPP-SFT while keeping CLIP frozen. The results are presented in Table~\ref{tab:ablation_freeze_vpt}. We can observe that making the global VPP learnable is beneficial. {This is because the pixel values of the axis-like image used for initialization may not be optimal for all images. By making the global VPP learnable, the model adjusts its parameters during training in a data-driven manner to better encode position-related cues. Although the global VPP remains fixed during inference, the learned representation provides generally useful spatial cues across inputs.}

\subsubsection{Variants of Global VPP Initialization}
\label{sec:var_gvpp}
\begin{table}[h]
\caption{The ablation study on variants of global VPP Initialization.}
\setlength\tabcolsep{2.2pt}
  \centering
  \renewcommand{\arraystretch}{1.3}
  \small  
  \begin{tabular}{lccccccccc}
  \toprule  
            \multirow{2}{*}{Type}&\multicolumn{3}{c}{RefCOCO} & \multicolumn{3}{c}{RefCOCO+}& \multicolumn{2}{c}{RefCOCOg}\\
            &val&testA&testB&val&testA&testB&val&test \\
            \midrule
            External-0.1 & 82.59 & 88.51 & 76.96 & 74.66 & 82.12 & 64.18 & 76.35 & 77.96 \\
            Internal-0.05 & 84.62 & 88.76 & 79.63 & 76.83 & 84.09 & 67.43 & 79.15 & 79.49 \\
            Cross-axis-0.1 & 84.68 & 88.49 & 78.88 & 77.55 & 83.60 & 66.56 & 78.64 & 79.96  \\
         \rowcolor{lightblue} Default & \textbf{85.29} &\textbf{89.96} &\textbf{80.06} &\textbf{77.68} & \textbf{84.75} & \textbf{67.62} & \textbf{79.80} & \textbf{80.36}  \\
     \bottomrule
     
  \end{tabular}
  \label{tab:ablation_different_GVPP}
\end{table}
{To evaluate how different initializations affect the model’s grounding performance, we conduct experiments under four distinct conditions, as shown in Fig.~\ref{fig:axis}:
    (1) using a unit scale of 0.1 with axes along the edges (Default in Table~\ref{tab:ablation_different_GVPP});
    (2) using a unit scale of 0.05 with axes along the edges (Internal-0.05 in Table~\ref{tab:ablation_different_GVPP});
    (3) using a unit scale of 0.1 with cross-shaped axes centered in the image (Cross-axis-0.1 in Table~\ref{tab:ablation_different_GVPP}); and
    (4) using a unit scale of 0.1 with externally padded axes along the edges (External-0.1 in Table~\ref{tab:ablation_different_GVPP}). Note that for conditions (3)) and (4), the binary mask must be removed.
We can draw several observations as follows. First, the results show that the default setting achieves better performance. Second, the External-0.1 strategy introduces a scale mismatch, as the actual image content occupies only 276×276 pixels within the 336×336 input. Since the pre-processing in our baseline model and the training data is based on the full 336×336 coordinate space, this mismatch leads to degraded performance. Lastly, the interval scale should be aligned with the internal coordinate system of the base MLLM, as different MLLMs may have varying internal representations.}

\subsubsection{Font Size of Global VPP Initialization}
    \definecolor{lightblue}{RGB}{240, 248, 255} 
        \begin{table}[h]
            \caption{{The ablation study on font size of the axis-like image for global VPP initialization.}}
            \setlength\tabcolsep{2.2pt}
              \centering
              \renewcommand{\arraystretch}{1.3}
              \small  
              \begin{tabular}{lccccccccc}
              \toprule  
                        \multirow{2}{*}{Font Size}&\multicolumn{3}{c}{RefCOCO} & \multicolumn{3}{c}{RefCOCO+}& \multicolumn{2}{c}{RefCOCOg}\\
                        &val&testA&testB&val&testA&testB&val&test \\
                        \midrule
                         5 & 85.05 & 89.94 & 79.69 & 77.47 & 84.03 & \textbf{67.70} & \textbf{79.92} & 79.73 \\
                        15 & 84.90 & 89.39 & 79.68 & 77.46 & 83.74 & 67.05 & 79.11 & 79.38 \\
                     \rowcolor{lightblue} Default (10) & \textbf{85.29} &\textbf{89.96} &\textbf{80.06} &\textbf{77.68} & \textbf{84.75} & 67.62 & 79.80 & \textbf{80.36}  \\
                 \bottomrule
                 
              \end{tabular}
              \label{tab:ablation_different_fontsize}
        \end{table}

{To further analyze the effect of initialization on model performance, we investigate the impact of font size used for coordinate digits in the axis-like image that initializes the global VPP. As shown in Table~\ref{tab:ablation_different_fontsize}, a moderate font size achieves the best trade-off. Both excessively small and large fonts result in slight performance degradation. Based on empirical observations, we adopt a font size of 10, which prevents digit overlap and ensures clear visual boundaries.}

\subsubsection{Type of Local VPP Generator}
\definecolor{lightblue}{RGB}{240, 248, 255} 
\begin{table}[t]
\caption{{Ablation study of different local VPP generators. A.DETR and V.DETR refer to Anchor-DETR and Vanilla-DETR. $^*$ denotes our adapted version (100 queries for A.DETR and 116 queries for YOLO-World).}}
\setlength\tabcolsep{2pt}
  \centering
  \renewcommand{\arraystretch}{1.3}
  \small  
  \begin{tabular}{lccccccccc}
  \toprule  
    \multirow{2}{*}{Type}&\multicolumn{3}{c}{RefCOCO} & \multicolumn{3}{c}{RefCOCO+}& \multicolumn{2}{c}{RefCOCOg}\\
    &val&testA&testB&val&testA&testB&val&test \\
    \midrule
        A. DETR & 84.21 & 89.22 & 79.21 & 76.00 & 84.19 & 67.01 & 78.76 & 79.38 \\
        A. DETR$^*$ & 85.10 & 89.04 & 79.55 & 77.57 & 84.33 & \textbf{68.62} & 79.22 & 79.56 \\
        YOLO-World$^*$ & 85.08 & \textbf{90.08} & 80.02 & 77.54 & \textbf{85.07} & 67.78 & 79.70 & 80.29 \\
        \rowcolor{lightblue} 
        V. DETR & \textbf{85.29} & 89.96 &\textbf{80.06} &\textbf{77.68} & 84.75 & 67.62 & \textbf{79.80} & \textbf{80.36}  \\
     \bottomrule
  \end{tabular}
   \label{tab:ablation_detr}
\end{table}
{We further investigate the effect of using different detectors to generate local VPP features. As shown in Table~\ref{tab:ablation_detr}, we compare Anchor-DETR (A.DETR)~\cite{anchor_detr}, Vanilla-DETR (V.DETR)~\cite{detr}, and the open-vocabulary detector YOLO-World~\cite{yolo_world}.}

{V.DETR consistently yields the strongest performance across all splits. Interestingly, reducing the number of object queries in A.DETR improves its results, suggesting that an excessive number of proposals may interfere with the LLM's spatial reasoning. Meanwhile, the YOLO-World variant—where we adapt its multi-scale features via downsampling and project them into the LLM space—also performs competitively, demonstrating the flexibility of our method to work with both closed-set and open-vocabulary detectors.}


{Moreover, as shown in our previous zero-shot evaluations on ReferIt and GSEval-BBox, our default V.DETR setup already exhibits strong zero-shot/open-vocabulary capabilities when integrated into the VPP framework. For these reasons, we opt to use V.DETR in our main experiments.}

\subsubsection{Fusion Strategy}
\definecolor{lightblue}{RGB}{240, 248, 255} 
\begin{table}[t]
\caption{The ablation study of the fusion strategy. \textit{C.A.1}, \textit{C.A.2} and \textit{Cat.} are the abbreviation for cross-attention-1, cross-attention-2, and concatenate operation respectively.}
\setlength\tabcolsep{3pt}
  \centering
  \renewcommand{\arraystretch}{1.3}
  \small  
  \begin{tabular}{lccccccccc}
  \toprule  
            \multirow{2}{*}{Strategy}&\multicolumn{3}{c}{RefCOCO} & \multicolumn{3}{c}{RefCOCO+}& \multicolumn{2}{c}{RefCOCOg}\\
            &val&testA&testB&val&testA&testB&val&test \\
            \midrule
            C.A.1 & 40.37 & 43.15 & 34.34 & 26.34 & 30.20 & 21.47 & 29.98 & 29.94 \\
            C.A.2 & 82.74 & 86.97 & 76.94 & 73.71 & 80.49 & 63.37 & 76.33 & 76.42 \\
         \rowcolor{lightblue} Cat. & \textbf{85.29} &\textbf{89.96} &\textbf{80.06} &\textbf{77.68} & \textbf{84.75} & \textbf{67.62} & \textbf{79.80} & \textbf{80.36}  \\
     \bottomrule
  \end{tabular}
  
   \label{tab:ablation_fusion_strategy}
\end{table}
Table~\ref{tab:ablation_fusion_strategy} presents our ablation study on fusion strategies for combining $F_{gp}'$ (CLIP visual features with global VPP) and $F_{lp}'$ (features from the local VPP generator) as outlined in Eq.~\ref{eq:fusion}. Here, \textit{C.A.1} stands for \textit{cross-attention-1}, which uses $F_{lp}'$ as the query and $F_{gp}'$ as the key in cross-attention, while \textit{C.A.2} reverses this, using $F_{gp}'$ as the query and $F_{lp}'$ as the key. \textit{Cat.} denotes direct concatenation. The cross-attention modules, trained with a learning rate of 2e-4, underperform relative to simple concatenation. This may be due to their training from scratch, which limits alignment with the LLM's pre-trained representations. Consequently, we adopt direct concatenation for our final model.

\subsubsection{Freeze or Unfreeze Visual Encoder}
\label{ablation:freeze}
\definecolor{lightblue}{RGB}{240, 248, 255} 
\begin{table}[t]
\caption{The ablation study examines whether to freeze or unfreeze the visual encoder.}
\setlength\tabcolsep{3pt}
  \centering
  \renewcommand{\arraystretch}{1.3}
  \small  
  \begin{tabular}{ccccccccc}
  \toprule  
            \multirow{2}{*}{Unfreeze}&\multicolumn{3}{c}{RefCOCO} & \multicolumn{3}{c}{RefCOCO+}& \multicolumn{2}{c}{RefCOCOg}\\
            &val&testA&testB&val&testA&testB&val&test \\
            \midrule
            \ding{55} & 85.29 & 89.96 & 80.06 & 77.68 & 84.75 & 67.62 & 79.80 & 80.36 \\
         \rowcolor{lightblue} \ding{51} & \textbf{86.48} &\textbf{91.25} &\textbf{81.59} &\textbf{79.75} & \textbf{86.22} & \textbf{70.38} & \textbf{81.14} & \textbf{82.14}  \\
     \bottomrule
  \end{tabular}
  
   \label{tab:ablation_freeze}
\end{table}
In Table~\ref{tab:ablation_freeze}, we report the results of freezing versus unfreezing the MLLM's visual encoder in our models. We observe that unfreezing the visual encoder significantly enhances performance on downstream visual grounding task. Since the introduced global VPP does not naturally exist in the MLLM's pretraining dataset, unfreezing the visual encoder allows it to better learn and adapt to this new form of input. This observation is also supported by recent research \cite{cambrian-1,eagle,llavanext}.

\subsubsection{Impact of VPP-Related Text Instructions}
\label{vpp_related text intruction}
\begin{table}[h]
\caption{{The Ablation Study on Levels of VPP-related Text Instructions.}}
    \setlength\tabcolsep{2pt}
    \centering
    \renewcommand{\arraystretch}{1.3}
    \small  
    \begin{tabular}{cccccccccc}
    \toprule  
    \multirow{2}{*}{System} &   \multirow{2}{*}{Sample}&\multicolumn{3}{c}{RefCOCO} & \multicolumn{3}{c}{RefCOCO+}& \multicolumn{2}{c}{RefCOCOg}\\
    &&val&testA&testB&val&testA&testB&val&test \\
    \midrule
    \ding{55} & \ding{55} & 85.17 & 89.37 & 79.67 & 77.02 & 84.58 & 67.25 & 78.92 & 79.59 \\
    \ding{51} & \ding{55} & 85.19 & \textbf{90.03} & 80.00 & \textbf{77.98} & 84.54 & \textbf{67.72} & 79.53 & 79.47 \\
    \rowcolor{lightblue} \ding{55} & \ding{51} & \textbf{85.29} & 89.96 &\textbf{80.06} & 77.68 & \textbf{84.75} & 67.62 & \textbf{79.80} & \textbf{80.36}  \\
    \bottomrule
    \end{tabular}
    \label{tab:ablation_instruction_presence}
\end{table}
{By default, each training sample in VPP-SFT includes a VPP-related text instruction at the sample level, i.e., \textit{Each image is accompanied by axes. If the question pertains to the bounding box coordinates, refer to the axes for the response.} To assess the impact of this instruction on model performance, we compare three variants, as reported in Table~\ref{tab:ablation_instruction_presence}. The first variant removes the instruction entirely during training and testing. The second places the instruction at the system level (i.e., system prompt) but omits it at the sample level. The third follows our default setting, where the instruction is included at the sample level.}

{Experimental results indicate that removing VPP-related text instructions at both the system and sample levels leads to a consistent drop in performance, particularly on semantically complex benchmarks such as RefCOCOg. This suggests that explicit textual guidance is essential for effectively leveraging the global VPP. While the performance difference between system-level and sample-level instruction is relatively small, system-level prompts introduce global constraints that may hinder the model's generalization to other tasks or instruction formats. To retain flexibility and task-specific adaptability, we adopt sample-level instruction as the default.}

\subsubsection{Hyperparameters}
\label{hyper}
In our proposed VPP method, the key hyperparameters are the trade-off parameter $\alpha$ and the width $w$ of the binary mask $M_w$ (Eq.~\ref{eq:gvpp}). Specifically, $\alpha$ controls the strength of the global VPP overlay, while $M_w$ defines its visible range. The impact of these hyperparameters on visual grounding performance is shown in Fig.~\ref{fig:abltion_alpha} and Fig.~\ref{fig:ablation_width}.

As shown in Fig.~\ref{fig:abltion_alpha}, performance is optimal when $\alpha$ is set to 0.95 across all datasets. A smaller $\alpha$ reduces the visibility of the MLLM's visual features, leading to lower performance. As $\alpha$ increases, allowing greater visibility of the original MLLM features, grounding performance improves. However, when $\alpha$ reaches 1 (no global VPP), performance declines, suggesting that excessive transparency negatively impacts the model. Thus, the global VPP should enhance rather than overpower the original visual features.

For the binary mask width $w$, as shown in Fig.~\ref{fig:ablation_width}. The default width of 30 (used for initializing axis-like images, as shown in Fig.~\ref{fig:framework}) yields the best performance. Both overly small ($w=10$) and overly large ($w=\text{None}$, no mask) values hinder performance. A small $w$ limits the visibility of the coordinate axes, reducing the effectiveness of the global VPP, while a large $w$ introduces excessive distortion, negatively impacting the visual features. Therefore, a balanced mask width is crucial for optimal performance.

\section{Qualitative Results}
\definecolor{Red}{RGB}{255,0,0}
\definecolor{Green}{RGB}{17,204,34}
\definecolor{Brown}{RGB}{190.0, 129.0,65.0}
\begin{figure*}[t]
    \centering
    \includegraphics[width=1\linewidth]{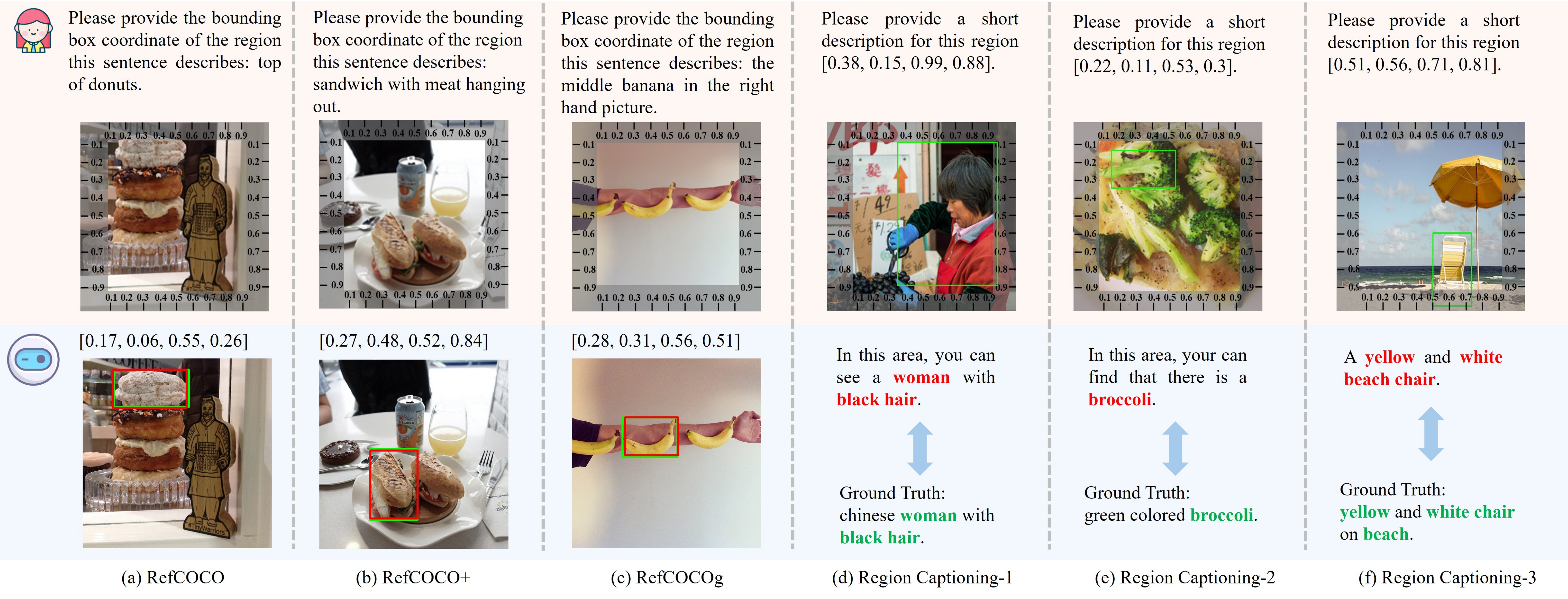}
    \caption{Qualitative results of our VPP-LLaVA on visual grounding and region captioning tasks. For brevity, some text instructions are omitted. \color{Green}{Green: ground truth}; \color{Red}{Red: ours}.}
    \label{fig:visualization}
\end{figure*}

\definecolor{Red}{RGB}{255,0,0}
\definecolor{Green}{RGB}{17,204,34}
\definecolor{Brown}{RGB}{190.0, 129.0,65.0}
\begin{figure*}[t]
    \centering
    \includegraphics[width=1\linewidth]{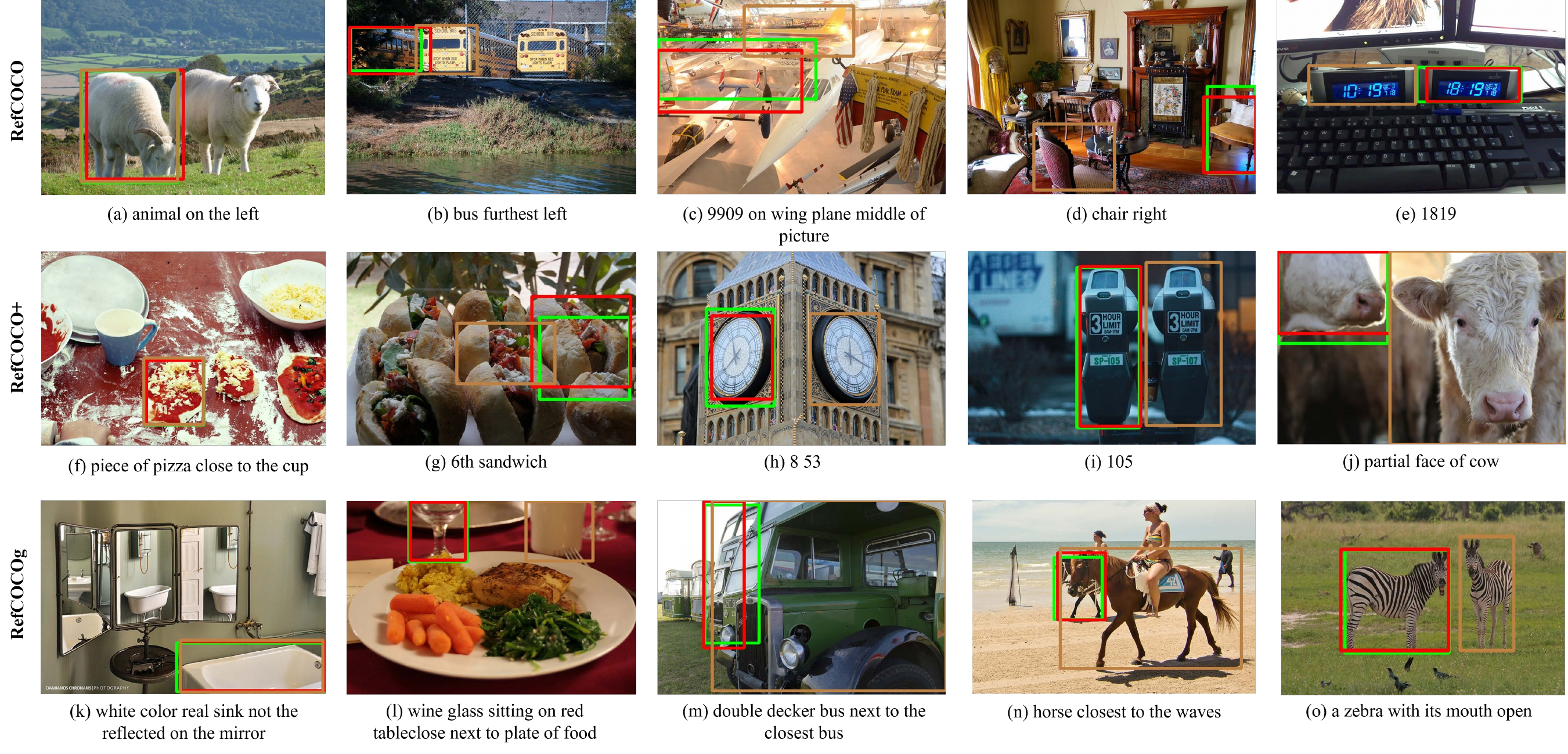}
    \caption{Qualitative visual grounding results of our VPP-LLaVA compared with the state-of-the-art method MiniGPT-v2 on RefCOCO dataset. For brevity, some text instructions are omitted. \color{Green}{Green: ground truth}; \color{Red}{Red: ours}; \color{Brown}{Brown: MiniGPT-v2}.}
    \label{fig:vpp-llava-tmm}
\end{figure*}

\definecolor{Red}{RGB}{255,0,0}
\definecolor{Green}{RGB}{17,204,34}
\definecolor{Purple}{RGB}{107.0, 41.0,205.0}
\begin{figure*}[t]
    \centering
    \includegraphics[width=1\linewidth]{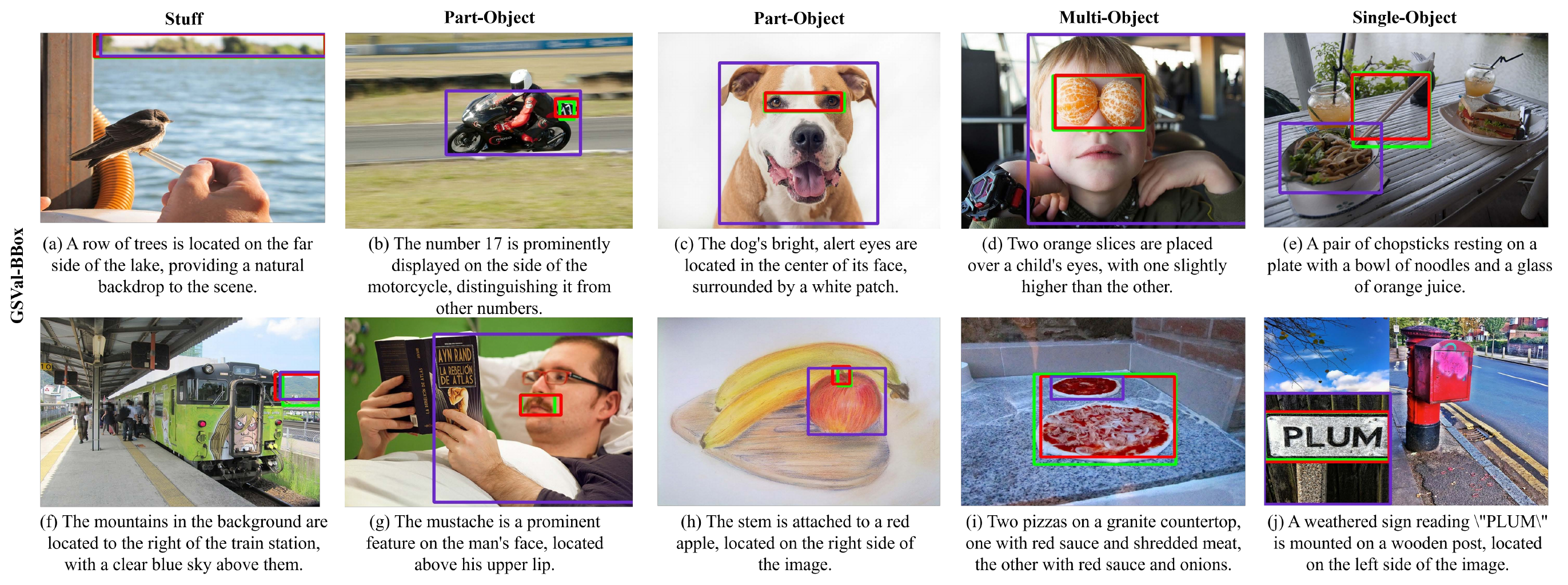}
    \caption{Qualitative visual grounding results of our VPP-LLaVA compared with the state-of-the-art method Qwen2.5-VL-7B on GSEval-BBox dataset with zero-shot settings. For brevity, some text instructions are omitted. \color{Green}{Green: ground truth}; \color{Red}{Red: ours}; \color{Purple}{Purple: Qwen2.5-VL-7B}.}
    \label{fig:vpp-llava-tmm}
\end{figure*}

\begin{figure*}[t]
    \centering
    \includegraphics[width=1\linewidth]{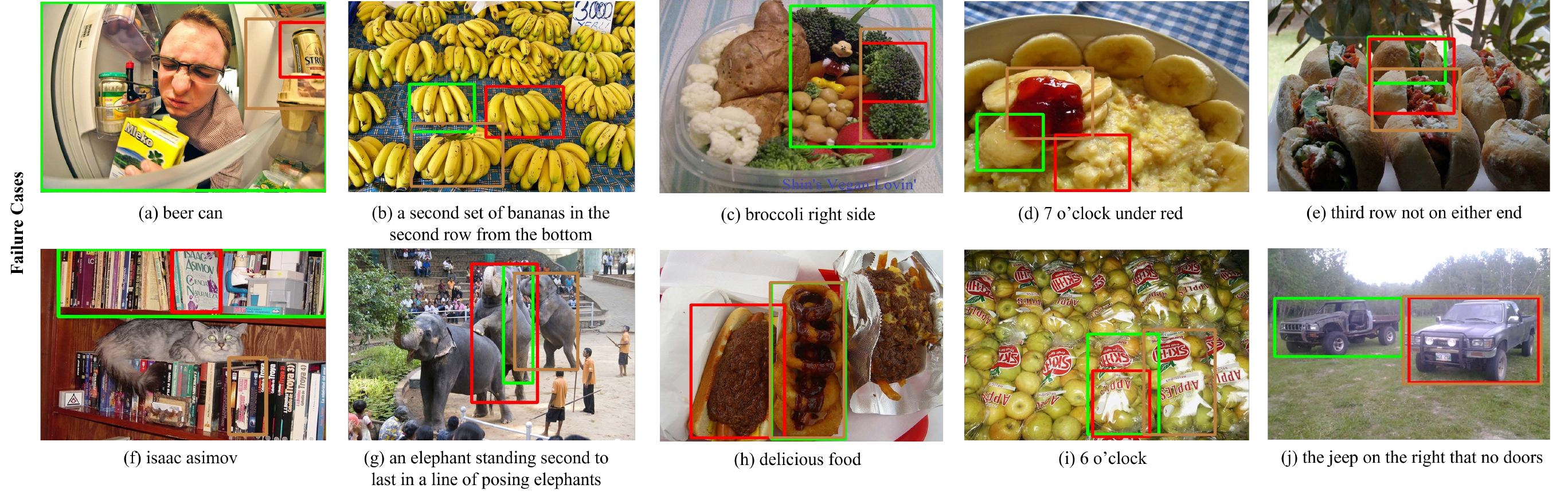}
    \caption{Some failure cases on visual grounding testing datasets. For brevity, some text instructions are omitted. \color{Green}{Green: ground truth}; \color{Red}{Red: ours}; \color{Brown}{Brown: MiniGPT-v2}.}
    \label{fig:failure}
    \vspace{-3mm}
\end{figure*}

\subsection{Qualitative Dialogue}
To fully understand the position-enhanced VPP-LLaVA for grounding tasks, we present our qualitative dialogue in Fig.~\ref{fig:visualization}. The predicted bounding box is marked in red, and the ground truth is marked in green. For brevity, some instruction prompts are omitted, and the full form of the conversation can be found in Fig.~\ref{fig:data_example}. Specifically, cases (a), (b), and (c) are three visual grounding examples from RefCOCO, RefCOCO+, and RefCOCOg, respectively. We can observe that our VPP-LLaVA demonstrates strong visual grounding capacity on these widely used benchmarks through the proposed global and local VPP. 

Furthermore, to demonstrate that our model retains satisfactory language capabilities after fine-tuning for visual grounding, we present two region captioning results on the RefCOCOg dataset in cases (d), (e) and (f). For ease of understanding, we mark the ground truth bounding boxes and the language phrases in these cases. We find that VPP-LLaVA accurately identifies the given regions and provides essentially correct captions. For example, in the final case of Fig.~\ref{fig:visualization}, our model accurately understands the given coordinates and outputs a suitable text response that includes the beach chair, along with its fine-grained attributes: yellow and white.

Overall, it is evident that the introduction of VPPs effectively assists the model in aligning spatial information in the images with coordinates, thereby enabling a better understanding of the images. 

\subsection{Qualitative Comparisons on RefCOCO/+/g}
Additional qualitative results for visual grounding are presented in Fig.~\ref{fig:vpp-llava-tmm}. Here, the green, red, and brown colors correspond to the bounding boxes of the ground truth, our VPP-LLaVA, and miniGPT-v2, respectively. For simplicity, we omit the \textit{7B} label in subsequent annotations.

Compared to the MiniGPT-v2 method, VPP-LLaVA demonstrates enhanced accuracy and reliability. For instance, in complex scenarios where the visibility of the target object is limited, as seen in Fig.~\ref{fig:vpp-llava-tmm} (b), (c), and (d), VPP-LLaVA accurately identifies the target object, while MiniGPT-v2 struggles. Similar situations can be observed in case (j) and case (l).

Leveraging the global VPP, our method offers superior global positional references when the query lacks explicit subject localization. This capability is particularly beneficial in scenarios where the absolute location of the query is not clearly indicated, allowing our approach to perform more effectively. Cases (e), (g), (h), and (i) illustrate this point clearly.

Furthermore, by incorporating local VPP, VPP-LLaVA offers enhanced object potential location and semantic information. This improvement enables the model to more easily distinguish between objects of similar categories. For instance, in case (m), VPP-LLaVA accurately identifies the double-decker bus, whereas the MiniGPT-v2 incorrectly locates a nearby bus. Similar situations can also be found in cases (n) and (o).

\subsection{Qualitative Comparisons on Zero-shot GSEval-BBox}
{To further evaluate the robustness of our VPP-LLaVA model in zero-shot scenarios, we present qualitative results on the GSEval-BBox dataset. As a comparison method, we include Qwen2.5-VL-7B, which ranks second in Table~\ref{tab:gseval} and represents one of the most advanced open-source MLLMs. The bounding boxes predicted by Qwen2.5-VL-7B are shown in purple. We illustrate several challenging referent localization cases across different levels of granularity.}

{From the results, we observe several key findings. In conjunction with Table~\ref{tab:gseval}, although our model shows a performance gap on stuff categories—likely due to limited data coverage and possible dataset bias—we still achieve competitive overall results, even under complex textual prompts. Notably, our model demonstrates remarkable performance on part-object cases. This is particularly evident in cases (b) and (h), where small object recognition poses a long-standing challenge in visual grounding, especially when the target is a part of a larger object and the prompt is linguistically complex. In these cases, our model produces highly accurate localizations. In contrast, even the state-of-the-art Qwen2.5-VL-7B, despite being trained on significantly more data, fails to precisely identify the part-object and instead predicts bounding boxes covering the entire main object—for example, the full motorcycle in case (b) and the whole apple in case (h). }

{Moreover, our model consistently handles both multi-object and single-object scenarios well on this dataset. While Qwen2.5-VL-7B also performs reasonably in some of these cases, a performance gap remains in favor of VPP-LLaVA. For instance, in case (i), our model accurately localizes two pizzas as instructed by the prompt, whereas Qwen2.5-VL-7B identifies only one. In case (e), Qwen2.5-VL-7B even fails to recognize the correct referent, mistakenly localizing an unrelated object, while our model identifies the intended target precisely.}

{These visualizations highlight the effectiveness of our VPP-LLaVA model, which integrates both global and local positional references. This combination is particularly beneficial in complex scenes where spatial understanding is crucial. The strong performance also stems from our high-quality VPP-SFT data—despite a relatively small training scale of 0.6M samples, our model achieves impressive zero-shot results, especially on challenging expressions and referent localization. These observations further demonstrate the robustness and efficiency of our approach.}

\subsection{Failure Cases Study}
To provide a comprehensive analysis of VPP-LLaVA, we present several failure cases from the involved datasets in Fig.~\ref{fig:failure}. For clarity, the \textit{7B} label is omitted in subsequent annotations.
When comparing with the ground truth and MiniGPT-v2, several key observations can be made:

First, in some instances, the ground truth annotations are not entirely accurate. For example, in cases (a) and (f), our method demonstrates its capability to provide a more precise location for the queried object than ground truth annotations. This demonstrates the effectiveness of our proposed VPPs, which allow the model to align image spatial information with coordinate details more accurately, thereby improving visual grounding performance.

Second, both VPP-LLaVA and MiniGPT-v2 may struggle to provide accurate bounding boxes for queries involving complex relational reasoning. Such scenarios remain challenging even for MLLMs with advanced reasoning capabilities. This limitation is evident in cases (b) and (g), where the relationships between objects are particularly intricate.

Third, ambiguous queries that could refer to multiple areas within an image pose significant challenges for both methods. This misalignment is particularly evident in cases (c) and (h), where the ambiguity of the query poses a significant challenge. The situation becomes even more complex when the subject is missing, as seen in cases (d) and (i), significantly increasing the difficulty of accurate visual grounding.

Lastly, for queries involving negations, such as in cases (e) and (j), both our method and MiniGPT-v2 may exhibit suboptimal performance. This is likely due to ongoing challenges in semantic understanding and reasoning related to negation for MLLMs, as negation requires deeper language comprehension and the ability to infer excluded information.

\section{Conclusion and Limitations}
In this paper, we propose VPP-LLaVA, an MLLM-based framework for visual grounding that incorporates the Visual Position Prompt (VPP). The global VPP initializes in an axis-like form and directly overlays onto the input image, providing a foundational position reference for the model. Additionally, the local VPP, serving as a local position reference, provides potential object locations and semantic information. Through these prompts, MLLMs can more accurately align image spatial information with coordinate details, thereby improving the visual grounding performance. Experiments indicate that, even when trained on limited data, our method outperforms state-of-the-art methods on widely used visual grounding benchmarks, demonstrating the superiority of our approach.

{The main limitations of this work are as follows. First, although our method achieves strong gains with limited training data, it still encounters challenges in handling complex relational reasoning and queries involving ambiguity or negation, which require deeper contextual understanding. Second, while the model shows strong performance and generalization in coordinate-related tasks, its capabilities on broader multimodal tasks remain underexplored and present opportunities for further enhancement. Future work may focus on incorporating more diverse and large-scale supervision or reinforcement learning methods to improve general MLLM abilities.}

\section{Acknowledgments}
This work was supported by National Natural Science Foundation of China (Grant No. 62425603), Basic Research Program of Jiangsu Province (Grant No. BK20240011), and Shanghai Artificial Intelligence Laboratory.

\bibliographystyle{IEEEtran}
\bibliography{reference}

\begin{thebibliography}{10}
\providecommand{\url}[1]{#1}
\csname url@samestyle\endcsname
\providecommand{\newblock}{\relax}
\providecommand{\bibinfo}[2]{#2}
\providecommand{\BIBentrySTDinterwordspacing}{\spaceskip=0pt\relax}
\providecommand{\BIBentryALTinterwordstretchfactor}{4}
\providecommand{\BIBentryALTinterwordspacing}{\spaceskip=\fontdimen2\font plus
\BIBentryALTinterwordstretchfactor\fontdimen3\font minus \fontdimen4\font\relax}
\providecommand{\BIBforeignlanguage}[2]{{%
\expandafter\ifx\csname l@#1\endcsname\relax
\typeout{** WARNING: IEEEtran.bst: No hyphenation pattern has been}%
\typeout{** loaded for the language `#1'. Using the pattern for}%
\typeout{** the default language instead.}%
\else
\language=\csname l@#1\endcsname
\fi
#2}}
\providecommand{\BIBdecl}{\relax}
\BIBdecl

\bibitem{tmm_4}
Y.~Li, B.~Hu, X.~Chen, L.~Ma, Y.~Xu, and M.~Zhang, ``Lmeye: An interactive perception network for large language models,'' \emph{IEEE Trans. Multimedia}, vol.~26, pp. 10\,952--10\,964, 2024.

\bibitem{llava}
H.~Liu, C.~Li, Q.~Wu, and Y.~J. Lee, ``Visual instruction tuning,'' in \emph{Proceedings of Advances in Neural Information Processing Systems}, vol.~36, 2024.

\bibitem{llava1.5}
H.~Liu, C.~Li, Y.~Li, and Y.~J. Lee, ``Improved baselines with visual instruction tuning,'' in \emph{Proceedings of the IEEE/CVF Conference on Computer Vision and Pattern Recognition}, 2024, pp. 26\,296--26\,306.

\bibitem{llavanext}
\BIBentryALTinterwordspacing
H.~Liu, C.~Li, Y.~Li, B.~Li, Y.~Zhang, S.~Shen, and Y.~J. Lee, ``Llava-next: Improved reasoning, ocr, and world knowledge,'' January 2024. [Online]. Available: \url{https://llava-vl.github.io/blog/2024-01-30-llava-next/}
\BIBentrySTDinterwordspacing

\bibitem{sun_1}
Y.~Sun, J.~Hao, K.~Zhu, J.-J. Liu, Y.~Zhao, X.~Li, G.~Zhang, Z.~Li, and J.~Wang, ``Descriptive caption enhancement with visual specialists for multi-modal perception,'' \emph{arXiv preprint arXiv:2412.14233}, 2025.

\bibitem{groma}
C.~Ma, Y.~Jiang, J.~Wu, Z.~Yuan, and X.~Qi, ``Groma: Localized visual tokenization for grounding multi-modal large language models,'' in \emph{Proceedings of European Conference on Computer Vision}, 2025, pp. 417--435.

\bibitem{ferret}
H.~You, H.~Zhang, Z.~Gan, X.~Du, B.~Zhang, Z.~Wang, L.~Cao, S.~Chang, and Y.~Yang, ``Ferret: Refer and ground anything anywhere at any granularity,'' in \emph{Proceedings of the International Conference on Learning Representations}, 2024.

\bibitem{Cong_1}
G.~Cong, L.~Li, Z.~Liu, Y.~Tu, W.~Qin, S.~Zhang, C.~Yan, W.~Wang, and B.~Jiang, ``{LS-GAN:} iterative language-based image manipulation via long and short term consistency reasoning,'' in \emph{Proceedings of ACM Int. Conf. Multimedia}, 2022, pp. 4496--4504.

\bibitem{tu_1}
Y.~Tu, L.~Li, L.~Su, Z.~Zha, and Q.~Huang, ``{SMART:} syntax-calibrated multi-aspect relation transformer for change captioning,'' \emph{{IEEE} Trans. Pattern Anal. Mach. Intell.}, vol.~46, no.~7, pp. 4926--4943, 2024.

\bibitem{bu2025error}
Y.~Bu, X.~Wu, Y.~Cai, Q.~Liu, T.~Wang, and Q.~Huang, ``Error-aware generative reasoning for zero-shot visual grounding,'' \emph{IEEE Trans. Multimedia}, 2025.

\bibitem{xie2025phrase}
M.~Xie, M.~Wang, H.~Li, Y.~Zhang, D.~Tao, and Z.~Yu, ``Phrase decoupling cross-modal hierarchical matching and progressive position correction for visual grounding,'' \emph{IEEE Trans. Multimedia}, 2025.

\bibitem{clip_vg}
L.~Xiao, X.~Yang, F.~Peng, M.~Yan, Y.~Wang, and C.~Xu, ``{CLIP-VG:} self-paced curriculum adapting of {CLIP} for visual grounding,'' \emph{IEEE Trans. Multimedia}, vol.~26, pp. 4334--4347, 2024.

\bibitem{minigptv2}
J.~Chen, D.~Zhu, X.~Shen, X.~Li, Z.~Liu, P.~Zhang, R.~Krishnamoorthi, V.~Chandra, Y.~Xiong, and M.~Elhoseiny, ``Minigpt-v2: Large language model as a unified interface for vision-language multi-task learning,'' \emph{arXiv preprint arXiv:2310.09478}, 2023.

\bibitem{DAC}
H.~Tang, Z.~Li, D.~Zhang, S.~He, and J.~Tang, ``Divide-and-conquer: confluent triple-flow network for {RGB-T} salient object detection,'' \emph{{IEEE} Trans. Pattern Anal. Mach. Intell.}, vol.~47, no.~3, pp. 1958--1974, 2025.

\bibitem{detr}
N.~Carion, F.~Massa, G.~Synnaeve, N.~Usunier, A.~Kirillov, and S.~Zagoruyko, ``End-to-end object detection with transformers,'' in \emph{Proceedings of European Conference on Computer Vision}, 2020, pp. 213--229.

\bibitem{anchor_detr}
Y.~Wang, X.~Zhang, T.~Yang, and J.~Sun, ``Anchor detr: Query design for transformer-based detector,'' in \emph{Proceedings of the AAAI conference on artificial intelligence}, vol.~36, no.~3, 2022, pp. 2567--2575.

\bibitem{ctnet}
Z.~Li, Y.~Sun, L.~Zhang, and J.~Tang, ``Ctnet: context-based tandem network for semantic segmentation,'' \emph{{IEEE} Trans. Pattern Anal. Mach. Intell.}, vol.~44, no.~12, pp. 9904--9917, 2022.

\bibitem{singular}
Y.~Sun, Q.~Chen, X.~He, J.~Wang, H.~Feng, J.~Han, E.~Ding, J.~Cheng, Z.~Li, and J.~Wang, ``Singular value fine-tuning: Few-shot segmentation requires few-parameters fine-tuning,'' in \emph{Proceedings of Advances in Neural Information Processing Systems}, vol.~35, 2022, pp. 37\,484--37\,496.

\bibitem{tgeo}
Y.~Zhan, Z.~Xiong, and Y.~Yuan, ``Rsvg: Exploring data and models for visual grounding on remote sensing data,'' \emph{{IEEE} Trans. Geosci. Remote. Sens.}, vol.~61, pp. 1--13, 2023.

\bibitem{tmm_7}
S.-F. Chen, J.-C. Chen, I.-H. Jhuo, and Y.-Y. Lin, ``Improving visual object tracking through visual prompting,'' \emph{IEEE Trans. Multimedia}, pp. 1--12, 2025.

\bibitem{shikra}
K.~Chen, Z.~Zhang, W.~Zeng, R.~Zhang, F.~Zhu, and R.~Zhao, ``Shikra: Unleashing multimodal llm's referential dialogue magic,'' \emph{arXiv preprint arXiv:2306.15195}, 2023.

\bibitem{cambrian-1}
P.~Tong, E.~Brown, P.~Wu, S.~Woo, A.~Iyer, S.~C. Akula, S.~Yang, J.~Yang, M.~Middepogu, Z.~Wang, X.~Pan, R.~Fergus, Y.~LeCun, and S.~Xie, ``Cambrian-1: {A} fully open, vision-centric exploration of multimodal llms,'' in \emph{Proceedings of Advances in Neural Information Processing Systems}, 2024.

\bibitem{gpt4roi}
S.~Zhang, P.~Sun, S.~Chen, M.~Xiao, W.~Shao, W.~Zhang, Y.~Liu, K.~Chen, and P.~Luo, ``Gpt4roi: Instruction tuning large language model on region-of-interest,'' in \emph{Proceedings of European Conference on Computer Vision}, 2025, pp. 52--70.

\bibitem{qwen-vl}
J.~Bai, S.~Bai, S.~Yang, S.~Wang, S.~Tan, P.~Wang, J.~Lin, C.~Zhou, and J.~Zhou, ``Qwen-vl: A frontier large vision-language model with versatile abilities,'' \emph{arXiv preprint arXiv:2308.12966}, 2023.

\bibitem{arcana}
Y.~Sun, H.~Zhang, Q.~Chen, X.~Zhang, N.~Sang, G.~Zhang, J.~Wang, and Z.~Li, ``Improving multi-modal large language model through boosting vision capabilities,'' \emph{arXiv preprint arXiv:2410.13733}, 2024.

\bibitem{groundingGPT}
Z.~Li, Q.~Xu, D.~Zhang, H.~Song, Y.~Cai, Q.~Qi, R.~Zhou, J.~Pan, Z.~Li, V.~Tu, Z.~Huang, and T.~Wang, ``Groundinggpt: Language enhanced multi-modal grounding model,'' in \emph{Proceedings of the Annual Meeting of the Association for Computational Linguistics}, 2024, pp. 6657--6678.

\bibitem{lisa}
X.~Lai, Z.~Tian, Y.~Chen, Y.~Li, Y.~Yuan, S.~Liu, and J.~Jia, ``Lisa: Reasoning segmentation via large language model,'' in \emph{Proceedings of the IEEE/CVF Conference on Computer Vision and Pattern Recognition}, 2024, pp. 9579--9589.

\bibitem{llava_grounding}
H.~Zhang, H.~Li, F.~Li, T.~Ren, X.~Zou, S.~Liu, S.~Huang, J.~Gao, C.~Li, J.~Yang \emph{et~al.}, ``Llava-grounding: Grounded visual chat with large multi-modal models,'' in \emph{Proceedings of European Conference on Computer Vision}, 2025, pp. 19--35.

\bibitem{dettoolchain}
Y.~Wu, Y.~Wang, S.~Tang, W.~Wu, T.~He, W.~Ouyang, P.~Torr, and J.~Wu, ``Dettoolchain: A new prompting paradigm to unleash detection ability of mllm,'' in \emph{Proceedings of European Conference on Computer Vision}, vol. 15090, 2024, pp. 164--182.

\bibitem{scaffolding}
X.~Lei, Z.~Yang, X.~Chen, P.~Li, and Y.~Liu, ``Scaffolding coordinates to promote vision-language coordination in large multi-modal models,'' in \emph{Proceedings of International Conference on Computational Linguistics}, 2025, pp. 2886--2903.

\bibitem{jiang2024global}
X.~Jiang, H.~Tang, and Z.~Li, ``Global meets local: Dual activation hashing network for large-scale fine-grained image retrieval,'' \emph{IEEE Trans. on Knowledge and Data Engineering}, 2024.

\bibitem{jiang2024dvf}
X.~Jiang, H.~Tang, R.~Yan, J.~Tang, and Z.~Li, ``Dvf: Advancing robust and accurate fine-grained image retrieval with retrieval guidelines,'' in \emph{Proceedings of the 32nd ACM International Conference on Multimedia}, 2024, pp. 2379--2388.

\bibitem{gseval}
R.~Hu, L.~Zhu, Y.~Zhang, T.~Cheng, L.~Liu, H.~Liu, L.~Ran, X.~Chen, W.~Liu, and X.~Wang, ``Groundingsuite: Measuring complex multi-granular pixel grounding,'' \emph{arXiv preprint arXiv:2503.10596}, 2025.

\bibitem{qwen2.5}
S.~Bai, K.~Chen, X.~Liu, J.~Wang, W.~Ge, S.~Song, K.~Dang, P.~Wang, S.~Wang, J.~Tang \emph{et~al.}, ``Qwen2.5-vl technical report,'' \emph{arXiv preprint arXiv:2502.13923}, 2025.

\bibitem{tcsvt_1}
C.~Wang, W.~Feng, S.~Lyu, G.~Cheng, X.~Li, B.~Liu, and Q.~Zhao, ``A masked reference token supervision-based iterative visual-language framework for robust visual grounding,'' \emph{{IEEE} Trans. Circuits Syst. Video Technol.}, vol.~35, no.~1, pp. 75--90, 2025.

\bibitem{transcp}
W.~Tang, L.~Li, X.~Liu, L.~Jin, J.~Tang, and Z.~Li, ``Context disentangling and prototype inheriting for robust visual grounding,'' \emph{IEEE Trans. Pattern Anal. Mach. Intell.}, vol.~46, no.~5, pp. 3213--3229, 2024.

\bibitem{EARN}
X.~Liu, L.~Li, S.~Wang, Z.-J. Zha, Z.~Li, Q.~Tian, and Q.~Huang, ``Entity-enhanced adaptive reconstruction network for weakly supervised referring expression grounding,'' \emph{IEEE Trans. Pattern Anal. Mach. Intell.}, vol.~45, no.~3, pp. 3003--3018, 2023.

\bibitem{mattnet}
L.~Yu, Z.~Lin, X.~Shen, J.~Yang, X.~Lu, M.~Bansal, and T.~L. Berg, ``Mattnet: Modular attention network for referring expression comprehension,'' in \emph{Proceedings of the IEEE/CVF Conference on Computer Vision and Pattern Recognition}, 2018, pp. 1307--1315.

\bibitem{2023cycleREC}
M.~Sun, J.~Xiao, E.~G. Lim, and Y.~Zhao, ``Cycle-free weakly referring expression grounding with self-paced learning,'' \emph{IEEE Trans. Multimedia}, vol.~25, pp. 1611--1621, 2023.

\bibitem{FAOA}
Z.~Yang, B.~Gong, L.~Wang, W.~Huang, D.~Yu, and J.~Luo, ``A fast and accurate one-stage approach to visual grounding,'' in \emph{Proceedings of the IEEE/CVF International Conference on Computer Vision}, 2019, pp. 4683--4693.

\bibitem{tip_1}
Y.~Liao, A.~Zhang, Z.~Chen, T.~Hui, and S.~Liu, ``Progressive language-customized visual feature learning for one-stage visual grounding,'' \emph{IEEE Trans. Image Process.}, vol.~31, pp. 4266--4277, 2022.

\bibitem{VLTVG}
L.~Yang, Y.~Xu, C.~Yuan, W.~Liu, B.~Li, and W.~Hu, ``Improving visual grounding with visual-linguistic verification and iterative reasoning,'' in \emph{Proceedings of the IEEE/CVF Conference on Computer Vision and Pattern Recognition}, 2022, pp. 9499--9508.

\bibitem{MDETR}
A.~Kamath, M.~Singh, Y.~LeCun, G.~Synnaeve, I.~Misra, and N.~Carion, ``Mdetr: Modulated eetection for end-to-end multi-modal understanding,'' in \emph{Proceedings of the IEEE/CVF International Conference on Computer Vision}, 2021, pp. 1760--1770.

\bibitem{ofa}
P.~Wang, A.~Yang, R.~Men, J.~Lin, S.~Bai, Z.~Li, J.~Ma, C.~Zhou, J.~Zhou, and H.~Yang, ``Ofa: Unifying architectures, tasks, and modalities through a simple sequence-to-sequence learning framework,'' in \emph{Proceedings of International Conference on Machine Learning}, 2022, pp. 23\,318--23\,340.

\bibitem{uninext}
B.~Yan, Y.~Jiang, J.~Wu, D.~Wang, P.~Luo, Z.~Yuan, and H.~Lu, ``Universal instance perception as object discovery and retrieval,'' in \emph{Proceedings of the IEEE/CVF Conference on Computer Vision and Pattern Recognition}, 2023, pp. 15\,325--15\,336.

\bibitem{Transvg}
J.~Deng, Z.~Yang, T.~Chen, W.~Zhou, and H.~Li, ``Transvg: End-to-end visual grounding with transformers,'' in \emph{Proceedings of the IEEE/CVF International Conference on Computer Vision}, 2021, pp. 1769--1779.

\bibitem{llama}
H.~Touvron, T.~Lavril, G.~Izacard, X.~Martinet, M.-A. Lachaux, T.~Lacroix, B.~Rozi{\`e}re, N.~Goyal, E.~Hambro, F.~Azhar \emph{et~al.}, ``Llama: Open and efficient foundation language models,'' \emph{arXiv preprint arXiv:2302.13971}, 2023.

\bibitem{vicuna}
\BIBentryALTinterwordspacing
W.-L. Chiang, Z.~Li, Z.~Lin, Y.~Sheng, Z.~Wu, H.~Zhang, L.~Zheng, S.~Zhuang, Y.~Zhuang, J.~E. Gonzalez, I.~Stoica, and E.~P. Xing, ``Vicuna: An open-source chatbot impressing gpt-4 with 90\%* chatgpt quality,'' March 2023. [Online]. Available: \url{https://lmsys.org/blog/2023-03-30-vicuna/}
\BIBentrySTDinterwordspacing

\bibitem{minigpt}
D.~Zhu, J.~Chen, X.~Shen, X.~Li, and M.~Elhoseiny, ``Minigpt-4: Enhancing vision-language understanding with advanced large language models,'' in \emph{Proceedings of the International Conference on Learning Representations}, 2024.

\bibitem{kosmos}
Z.~Peng, W.~Wang, L.~Dong, Y.~Hao, S.~Huang, S.~Ma, and F.~Wei, ``Kosmos-2: Grounding multimodal large language models to the world,'' \emph{arXiv preprint arXiv:2306.14824}, 2023.

\bibitem{pink}
S.~Xuan, Q.~Guo, M.~Yang, and S.~Zhang, ``Pink: Unveiling the power of referential comprehension for multi-modal llms,'' in \emph{Proceedings of the IEEE/CVF Conference on Computer Vision and Pattern Recognition}, 2024, pp. 13\,838--13\,848.

\bibitem{glamm}
H.~Rasheed, M.~Maaz, S.~Shaji, A.~Shaker, S.~Khan, H.~Cholakkal, R.~M. Anwer, E.~Xing, M.-H. Yang, and F.~S. Khan, ``Glamm: Pixel grounding large multi-modal model,'' in \emph{Proceedings of the IEEE/CVF Conference on Computer Vision and Pattern Recognition}, 2024, pp. 13\,009--13\,018.

\bibitem{interactive_seg_1}
K.~Sofiiuk, I.~Petrov, O.~Barinova, and A.~Konushin, ``f-brs: Rethinking backpropagating refinement for interactive segmentation,'' in \emph{Proceedings of the IEEE/CVF Conference on Computer Vision and Pattern Recognition}, 2020, pp. 8623--8632.

\bibitem{tnnls_1}
Y.~Long, J.~Han, R.~Huang, H.~Xu, Y.~Zhu, C.~Xu, and X.~Liang, ``Fine-grained visual–text prompt-driven self-training for open-vocabulary object detection,'' \emph{{IEEE} Trans. Neural Networks Learn. Syst.}, vol.~35, no.~11, pp. 16\,277--16\,287, 2024.

\bibitem{jiang2024delving}
X.~Jiang, H.~Tang, J.~Gao, X.~Du, S.~He, and Z.~Li, ``Delving into multimodal prompting for fine-grained visual classification,'' in \emph{Proceedings of the AAAI conference on artificial intelligence}, vol.~38, no.~3, 2024, pp. 2570--2578.

\bibitem{visual_incontext}
Y.~Sun, Q.~Chen, J.~Wang, J.~Wang, and Z.~Li, ``Exploring effective factors for improving visual in-context learning,'' \emph{IEEE Trans. Image Process.}, vol.~34, pp. 2147--2160, 2025.

\bibitem{tcsvt_2}
X.~Liu, J.~Wu, W.~Yang, X.~Zhou, and T.~Zhang, ``Multi-modal attribute prompting for vision-language models,'' \emph{{IEEE} Trans. Circuits Syst. Video Technol.}, vol.~34, no.~11, pp. 11\,579--11\,591, 2024.

\bibitem{maple}
M.~U. Khattak, H.~Rasheed, M.~Maaz, S.~Khan, and F.~S. Khan, ``Maple: Multi-modal prompt learning,'' in \emph{Proceedings of the IEEE/CVF Conference on Computer Vision and Pattern Recognition}, 2023, pp. 19\,113--19\,122.

\bibitem{vpt}
M.~Jia, L.~Tang, B.-C. Chen, C.~Cardie, S.~Belongie, B.~Hariharan, and S.-N. Lim, ``Visual prompt tuning,'' in \emph{Proceedings of European Conference on Computer Vision}, 2022, pp. 709--727.

\bibitem{cmpa}
X.~Liu, W.~Tang, J.~Lu, R.~Zhao, Z.~Guo, and F.~Tan, ``Deeply coupled cross-modal prompt learning,'' in \emph{Findings of the Association for Computational Linguistics}, 2023, pp. 7957--7970.

\bibitem{pevl}
Y.~Yao, Q.~Chen, A.~Zhang, W.~Ji, Z.~Liu, T.-S. Chua, and M.~Sun, ``Pevl: Position-enhanced pre-training and prompt tuning for vision-language models,'' in \emph{Proceedings of Conference on Empirical Methods in Natural Language Processing}, 2022, pp. 11\,104--11\,117.

\bibitem{CPT}
Y.~Yao, A.~Zhang, Z.~Zhang, Z.~Liu, T.~Chua, and M.~Sun, ``Cpt: Colorful prompt tuning for pre-trained vision-language models,'' \emph{AI Open}, vol.~5, pp. 30--38, 2024.

\bibitem{sam}
A.~Kirillov, E.~Mintun, N.~Ravi, H.~Mao, C.~Rolland, L.~Gustafson, T.~Xiao, S.~Whitehead, A.~C. Berg, W.-Y. Lo \emph{et~al.}, ``Segment anything,'' in \emph{Proceedings of the IEEE/CVF International Conference on Computer Vision}, 2023, pp. 4015--4026.

\bibitem{vrp_sam}
Y.~Sun, J.~Chen, S.~Zhang, X.~Zhang, Q.~Chen, G.~Zhang, E.~Ding, J.~Wang, and Z.~Li, ``Vrp-sam: Sam with visual reference prompt,'' in \emph{Proceedings of the IEEE/CVF Conference on Computer Vision and Pattern Recognition}, 2024.

\bibitem{fine_grained_vp}
L.~Yang, Y.~Wang, X.~Li, X.~Wang, and J.~Yang, ``Fine-grained visual prompting,'' in \emph{Proceedings of Advances in Neural Information Processing Systems}, vol.~36, 2024.

\bibitem{vip-llava}
M.~Cai, H.~Liu, S.~K. Mustikovela, G.~P. Meyer, Y.~Chai, D.~Park, and Y.~J. Lee, ``Vip-llava: Making large multi-modal models understand arbitrary visual prompts,'' in \emph{Proceedings of the IEEE/CVF Conference on Computer Vision and Pattern Recognition}, 2024, pp. 12\,914--12\,923.

\bibitem{tvp}
Y.~Zhang, Y.~Dong, S.~Zhang, T.~Min, H.~Su, and J.~Zhu, ``Exploring the transferability of visual prompting for multi-modal large language models,'' in \emph{Proceedings of the IEEE/CVF Conference on Computer Vision and Pattern Recognition}, 2024, pp. 26\,562--26\,572.

\bibitem{rethink_som}
Y.~Lin, Y.~Li, D.~Chen, W.~Xu, R.~Clark, P.~Torr, and L.~Yuan, ``Rethinking visual prompting for multi-modal large language models with external knowledge,'' \emph{arXiv preprint arXiv:2407.04681}, 2024.

\bibitem{wang2024contextdet}
N.~Wang, Y.~Xiao, X.~Peng, X.~Chang, X.~Wang, and D.~Fang, ``Contextdet: Temporal action detection with adaptive context aggregation,'' \emph{arXiv preprint arXiv:2410.15279}, 2024.

\bibitem{chatterbox}
P.~Mavridis, O.~Huang, S.~Qiu, U.~Gadiraju, and A.~Bozzon, ``Chatterbox: Conversational interfaces for microtask crowdsourcing,'' in \emph{Proceedings of ACM Conference on User Modeling, Adaptation and Personalization}, 2019, pp. 243--251.

\bibitem{genixer}
H.~H. Zhao, P.~Zhou, and M.~Z. Shou, ``{GENIXER:} empowering multimodal large language model as a powerful data generator,'' in \emph{Proceedings of European Conference on Computer Vision}, vol. 15081, 2024, pp. 129--147.

\bibitem{mscoco}
T.-Y. Lin, M.~Maire, S.~Belongie, J.~Hays, P.~Perona, D.~Ramanan, P.~Doll{\'a}r, and C.~L. Zitnick, ``Microsoft coco: Common objects in context,'' in \emph{Proceedings of European Conference on Computer Vision}, 2014, pp. 740--755.

\bibitem{refcoco_google}
J.~Mao, J.~Huang, A.~Toshev, O.~Camburu, A.~L. Yuille, and K.~Murphy, ``Generation and comprehension of unambiguous object descriptions,'' in \emph{Proceedings of the IEEE/CVF International Conference on Computer Vision}, 2016, pp. 11--20.

\bibitem{refcoco_umd}
V.~K. Nagaraja, V.~I. Morariu, and L.~S. Davis, ``Modeling context between objects for referring expression understanding,'' in \emph{Proceedings of European Conference on Computer Vision}, 2016, pp. 792--807.

\bibitem{lion}
G.~Chen, L.~Shen, R.~Shao, X.~Deng, and L.~Nie, ``Lion: Empowering multimodal large language model with dual-level visual knowledge,'' in \emph{Proceedings of the IEEE/CVF Conference on Computer Vision and Pattern Recognition}, 2024, pp. 26\,540--26\,550.

\bibitem{eagle}
M.~Shi, F.~Liu, S.~Wang, S.~Liao, S.~Radhakrishnan, D.-A. Huang, H.~Yin, K.~Sapra, Y.~Yacoob, H.~Shi \emph{et~al.}, ``Eagle: Exploring the design space for multimodal llms with mixture of encoders,'' \emph{arXiv preprint arXiv:2408.15998}, 2024.

\bibitem{vit_big}
X.~Zhai, A.~Kolesnikov, N.~Houlsby, and L.~Beyer, ``Scaling vision transformers,'' in \emph{Proceedings of the IEEE/CVF conference on computer vision and pattern recognition}, 2022, pp. 12\,104--12\,113.

\bibitem{eva}
Y.~Fang, W.~Wang, B.~Xie, Q.~Sun, L.~Wu, X.~Wang, T.~Huang, X.~Wang, and Y.~Cao, ``Eva: Exploring the limits of masked visual representation learning at scale,'' in \emph{Proceedings of the IEEE/CVF Conference on Computer Vision and Pattern Recognition}, 2023, pp. 19\,358--19\,369.

\bibitem{griffon_v2}
Y.~Zhan, Y.~Zhu, H.~Zhao, F.~Yang, M.~Tang, and J.~Wang, ``Griffon v2: Advancing multi-modal perception with high-resolution scaling and visual-language co-referring,'' in \emph{Proceedings of European Conference on Computer Vision}, 2024.

\bibitem{internvl2.5}
Z.~Chen, W.~Wang, Y.~Cao, Y.~Liu, Z.~Gao, E.~Cui, J.~Zhu, S.~Ye, H.~Tian, Z.~Liu \emph{et~al.}, ``Expanding performance boundaries of open-source multimodal models with model, data, and test-time scaling,'' \emph{arXiv preprint arXiv:2412.05271}, 2024.

\bibitem{deepseekvl2}
Z.~Wu, X.~Chen, Z.~Pan, X.~Liu, W.~Liu, D.~Dai, H.~Gao, Y.~Ma, C.~Wu, B.~Wang \emph{et~al.}, ``Deepseek-vl2: Mixture-of-experts vision-language models for advanced multimodal understanding,'' \emph{arXiv preprint arXiv:2412.10302}, 2024.

\bibitem{mistral}
mistralai, ``Mistral-small-3.1-24b-instruct-2503,'' \url{https://huggingface.co/mistralai/Mistral-Small-3.1-24B-Instruct-2503}, 2024.

\bibitem{yolo_world}
T.~Cheng, L.~Song, Y.~Ge, W.~Liu, X.~Wang, and Y.~Shan, ``Yolo-world: Real-time open-vocabulary object detection,'' in \emph{Proceedings of the IEEE/CVF Conference on Computer Vision and Pattern Recognition}, 2024, pp. 16\,901--16\,911.

\end{thebibliography}

\end{document}